%% file: main.tex
\documentclass[runningheads]{llncs}
\usepackage{graphicx}
\usepackage{comment}
\usepackage{amsmath,amssymb} 
\usepackage{color}
\usepackage{mwe}
\usepackage{subfig}

\usepackage{gensymb}
\usepackage{hyperref}

\input{00_macros.tex}

\usepackage[width=122mm,left=12mm,paperwidth=146mm,height=193mm,top=12mm,paperheight=217mm]{geometry}
\begin{document}

\pagestyle{headings}
\mainmatter
\def\ECCVSubNumber{1535}  

\title{Can You Read Me Now? \\ Content Aware Rectification using Angle Supervision} 
\titlerunning{Content Aware Rectification using Angle Supervision}  
\author{Amir Markovitz* \and
Inbal Lavi* \and
Or Perel \and
Shai Mazor \and
Roee Litman
}
\authorrunning{A. Markovitz et al.}  
\institute{Amazon Web Services \\ 
\email{\{amirmak, ilavi, orperel, smazor, rlit\}@amazon.com}
}
\maketitle
\input{10_abstract.tex}

\input{20_introduction.tex}
\input{30_background.tex}

\input{40_method.tex}
\input{50_experiments.tex}

\input{60_conclusion.tex}

\clearpage

\input{main.bbl}
\clearpage
\appendix
\input{supp/100_supp_intro}
\input{supp/110_supp_docrect}
\pagebreak
\input{supp/120_supp_dataset}
\input{supp/130_supp_cart2polar}
\input{supp/135_supp_text_seg}
\input{supp/140_supp_ssim}
\input{supp/150_supp_vis_results}

\end{document}

%% file: 00_macros.tex
\usepackage{overpic}
\usepackage{multirow}

\newcommand{\etal}{{et al}.\@ }

\DeclareMathOperator{\atantwo}{atan2}
\DeclareMathOperator{\arctantwo}{arctan2}

\makeatletter
\renewcommand{\paragraph}{%
  \@startsection{paragraph}{4}%
  {\z@}{0.80ex \@plus 1ex \@minus .2ex}{-1em}%
  {\normalfont\normalsize\bfseries}%
}
\makeatother

\makeatletter
\def\blfootnote{\gdef\@thefnmark{}\@footnotetext}
\makeatother

\ifdefined\ShowNotes
  \newcommand{\colornote}[3]{{\color{#1}\bf{#2: #3}\normalfont}}
\else
  \newcommand{\colornote}[3]{}
\fi

%% file: 10_abstract.tex
\begin{abstract}
 
The ubiquity of smartphone cameras has led to more and more documents being captured by cameras rather than scanned.
Unlike flatbed scanners, photographed documents are often folded and crumpled, resulting in large local variance in text structure. 
The problem of document rectification is fundamental to the Optical Character Recognition (OCR) process on documents, and its ability to overcome geometric distortions significantly affects recognition accuracy.
Despite the great progress in recent OCR systems, most still rely on a pre-process that ensures the text lines are straight and axis aligned.
Recent works have tackled the problem of rectifying document images taken in-the-wild using various supervision signals and alignment means. However, they focused on global features that can be extracted from the document's boundaries, ignoring various signals that could be obtained from the document's content.

We present CREASE: Content Aware Rectification using Angle Supervision, the first learned method for document rectification that relies on the document's content, the location of the words and specifically their orientation, as hints to assist in the rectification process. We utilize a novel pixel-wise angle regression approach and a curvature estimation side-task for optimizing our rectification model.
Our method surpasses previous approaches in terms of OCR accuracy, geometric error and visual similarity.

\end{abstract}

%% file: 20_introduction.tex
\section{Introduction}

Documents are a common way to share information and record transactions between people. 
In order to digitize mass amounts of printed documents, the hard copies are scanned and text is extracted automatically by Optical Character Recognition (OCR) systems, such as \cite{textract,gcp_ocr}.
In the past, most documents were scanned in flatbed scanners. However, the past few years have seen a rise in the use of smartphones, and with it the use of the smartphone camera as a document scanner. 
Camera captured documents such as receipts are often folded, curved, or crumpled, and vary greatly in camera angles, lighting and texture conditions. This makes the OCR task much more challenging compared to scanned images.\blfootnote{* - Equal Contribution}

Recent OCR methods have had great success in recognizing text in very challenging scenarios. One example is scene text recognition \cite{baek2019STRcomparisons,lyu2018masktextspotter} which aims to recognize text in natural images. The text is often sparse, and may also be rotated or curved.
Another scenario is retrieving the content of a document with \emph{dense} text, that poses the challenge of detecting and recognizing many words that are closely located. 

\input{figures/teaser/teaser}
While recognizing dense text, and similarly detecting sparse curved text, had been studied thoroughly, the combined problem of both dense and warped text detection and recognition has received significantly less attention. 
Many text detectors assume axis-aligned text and struggle with deformed lines~\cite{grning2018twostage,yousef2020origaminet}, while text recognition systems struggle with fine deformations on the character level.
Taking this into account, a line of works proposed to rectify the document as a pre-process to the recognition phase.
Recent methods harnessed the power of deep learning to solve this task \cite{das2019dewarpnet,ma2018docunet}, but put more emphasis on the page boundaries and less emphasis on the contents of the document.

In this paper, we present CREASE: \emph{Content Aware Rectification using Angle Supervision}.
This method performs document rectification by relying on both global and local hints, with an emphasis on content. 
Our method predicts the 3D structure globally while simultaneously optimizing for the local structure of both the text orientation, the location of folds and creases, and the output backward map.
CREASE provides results that are superior in readability, similarity and geometric reconstruction. 

CREASE predicts the mapping of a warped document image to its ``flatbed'' version. First, we estimate the 3D structure of the input document. 
Then, we transform this estimation into a mapping, specifically, a backward mapping. Finally, the mapping is used to resample the warped image into the flattened form. A general overview of CREASE is given in Figure~\ref{fig:teaser}. Our contributions are as follows:
\begin{enumerate}
    \item We present a per-pixel angle regression loss that complements the 3D structure estimation by optimizing different aspects of the rectification process.
    \item We present a curvature estimation task, which predicts the lines along which the document is crumpled or folded, complementing the per-pixel angle regression loss by emphasizing its discontinuities.
    \item The losses are learned as side tasks, focusing on the areas of the document that contain strong signals regarding the text orientation, and are optimized alongside the 3D structure estimation in an end-to-end optimization process.
    \item We reduce the relative OCR error in a challenging warped document dataset by $20.2 \%$ and the relative geometric error by $14.1 \%$, compared to the state-of-the-art method. 
\end{enumerate}
    
We train CREASE using synthetic data, which provides us with intricate details regarding each document in our training set without requiring manual annotation: the ground-truth transformation for every pixel, the 3D coordinates, angles, curvature values for every pixel, and the text segmentation mask.

We present visual and quantitative results and comparisons on both synthetic and real evaluation datasets. We also present a detailed study of the contribution of each individual model component.

%% file: figures/teaser/teaser.tex
\begin{figure*}[t]
    \centering
    \includegraphics[width=0.95\textwidth]{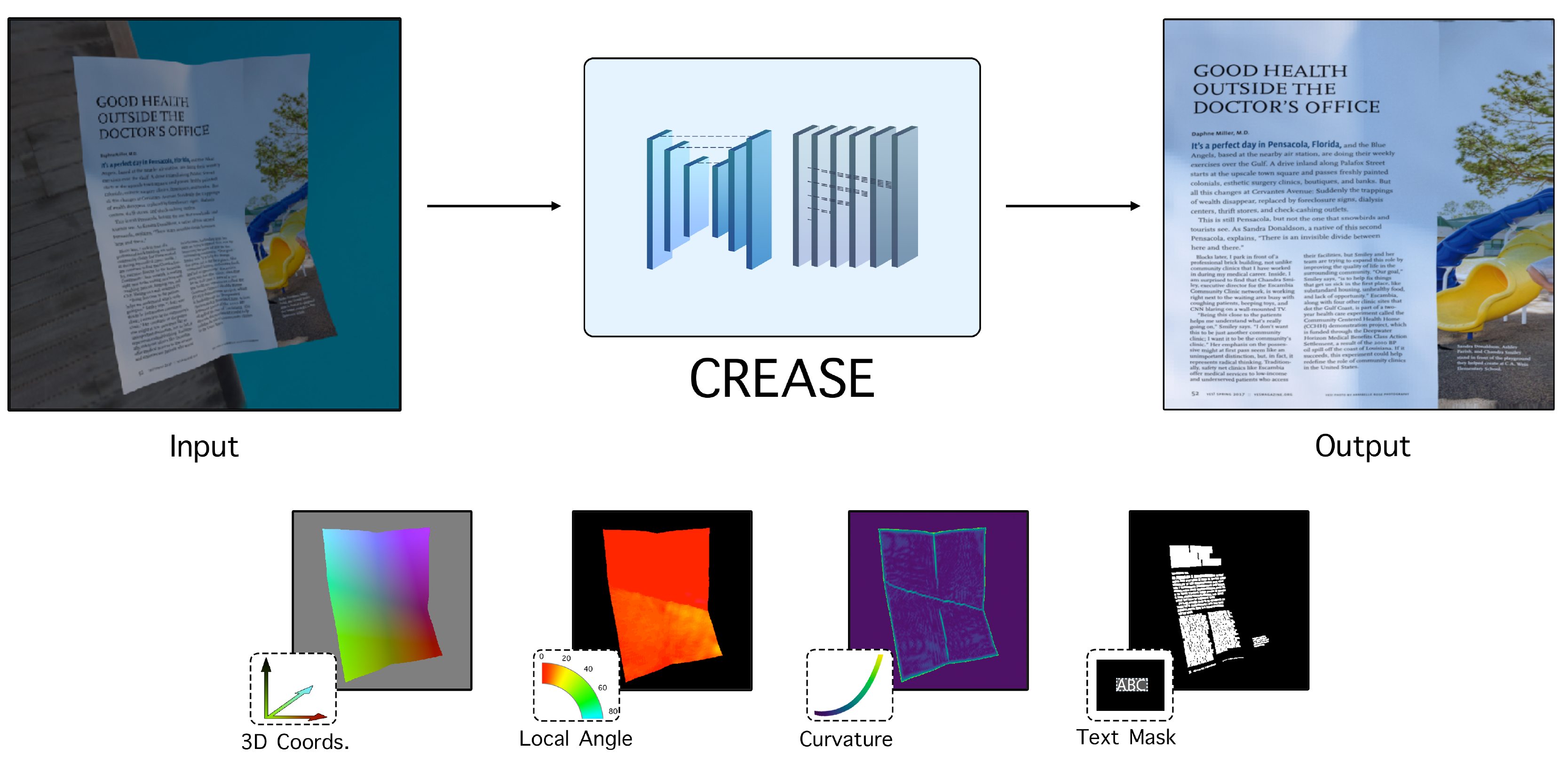}\\

    \caption{\label{fig:teaser}
        {\bf Overview of CREASE.}
        CREASE is a document rectification method that learns content based signals in addition to the document's 3D structure in order to estimate a transformation used for rectifying the image. On the left is a synthetically generated input image, and on the right is the image rectified using the transformation that CREASE predicted. The bottom images are the supervision signals, from left to right: The 3D coordinates of the warped document, the angle deformation map, the curvature map, and the text mask.
    }
    
\end{figure*}

%% file: 30_background.tex
\section{Background} \label{background}

Many works have addressed the problem of extracting text from documents captured in challenging scenarios. These works have focused on different elements that improve OCR accuracy, such as illumination and noise correction \cite{li2019docrect,Ramanna2019deeplearning}, resolution enhancement \cite{zheng2014real}, and document rectification~\cite{das2017fold,ma2018docunet,das2019dewarpnet}. This work focuses on the last problem of rectifying a warped document from a single image by prediction of the 3D model.

Early document rectification methods used hand crafted features to detect the structure of a document.
These methods usually made strong assumptions on the deformation process such as smoothness \cite{huang2015textline,burden2019rectification} and folding structure \cite{das2017fold}. 
Several works utilized special equipment to capture the 3D model of the document~\cite{Brown2004ImageRO}, or reconstructed the 3D model from multi-view images \cite{You2016MultiviewRO}.

More recently, approaches such as \cite{ma2018docunet,das2019dewarpnet,li2019docrect} have used deep learning to rectify single-image, camera-captured documents, and were designed to solve the rectification problem by directly predicting the document warp. 
These works placed the focus on the aforementioned warp, and gave no special treatment to the content of the document, i.e., the data we wish to ultimately recognize.

The first in this line of works is DocUnet \cite{ma2018docunet}, which used a stacked hourglass architecture to predict the original 2D coordinates of each pixel in the warped document. 
This prediction, in essence, gives the forward mapping from the rectified image to the warped one, which can be inverted to get the final result. 

A followup work presented DewarpNet \cite{das2019dewarpnet}, which added three learned post-processing components for calculating the backward map, the surface normals and the shading, each as a separate hourglass network. 
Additionally, the original 2D forward map of \cite{ma2018docunet} was replaced by a prediction of its 3D counterpart, and the stacked hourglass was substituted by a single hourglass network for this prediction. 
This method mostly relied on the document boundary, and did not explicitly address the document's content in the rectification module.

Another work by Li \etal \cite{li2019docrect} focused mainly on uneven background illumination, however it did this by predicting the document warp. 
This work computed a forward map in a method similar to \cite{ma2018docunet}, but divided the prediction into three phases. 
First, a local, patch-based network predicted the gradients of the forward map. 
Then, a graph-cut model stitched these patch-predictions into a global warp. 
Finally, the un-warped image underwent an illumination correction. 
The local and global level prediction allowed the warp estimation to take into account both the document boundaries and areas in the center of the document, but this pipeline required an expensive patch-stitching process and worked best on input documents with minimal background.
Additionally, this method was not end-to-end trainable and did not take into account the document content.

One of the key points in our approach is the importance of predicting text orientation in the warped image.
The notion of text angle prediction was previously explored in scene text recognition at the word (or, object) level \cite{zhou2017east,ma2017arbitraryoriented,liu2018fots}, as opposed to our pixel-level approach.

The EAST text detector by Zhou \etal \cite{zhou2017east} and FOTS detector by Liu \etal \cite{liu2018fots} both predicted the angle for each word detection candidate in conjunction with other parameters, like bounding box size and quadrangle coordinates.
Ma \etal \cite{ma2017arbitraryoriented} extended upon the Faster-RCNN \cite{ren2015faster} architecture by adding rotated anchors to accommodate for arbitrarily oriented text.
It is important to stress that scene text methods deal with a sparse set of words, and moreover, each word is rectified separately.
Documents, on the other hand, benefit more from a rectification process before word localization, due to the denser text but also due to a stronger prior on the structure thereof.
To make use of this prior, CREASE applies an angle regression loss at the pixel level, focused on the salient text areas in the document, and optimized in an end-to-end manner over the predicted backward map.

%% file: 40_method.tex
\section{Method}

We design CREASE to exploit a document's content and geometry on both local and global levels.
CREASE addresses different aspects of the input, such as global structure, creases and fold lines, and per-pixel angular deformation. This allows it to capture an accurate mapping for the entire document globally, and for fine-grained features such as characters and words locally (without detecting them explicitly).

First, we present the general architecture of our model (sub-section~\ref{method:arch}). Next, we present the properties of documents that CREASE relies on: the flow field angles (sub-section~\ref{method:angle}), and the curvature estimation (sub-section~\ref{method:crease}).
Finally, we present the optimization objective tying the various signals together (sub-section~\ref{method:optimization}).

\input{./figures/arch/arch.tex}
\subsection{Architecture} \label{method:arch}

CREASE is comprised of a two-stage network, illustrated in Figure \ref{fig:arch}. The first stage is used for estimating the location of each pixel in a normalized 3D coordinate system, the warp field angle values, and the curvature in each pixel. The 3D estimation module is followed by a backward mapping network. This network transforms the estimated 3D coordinate image into a backward map that can be used for rectifying the input image.

\subsubsection{3D Estimation.} The first stage provides per-pixel estimation for 3D coordinates (based on \cite{das2019dewarpnet}) along with the angle and curvature outputs, used as side-tasks. A Unet~\cite{ronneberger2015u} based architecture is used for mapping an input image into the angle, curvature and 3D coordinate maps. The three maps are used for supervision, and the 3D coordinate map also functions as the input for the backward mapping stage.

\subsubsection{Backward Mapping.}
The second network stage transforms the 3D coordinate map outputted from the first stage to the backward mapping of the image. This mapping describes the transformation from the warped image to the rectified result. In other words, the backward map determines for every pixel in the rectified (output) image domain, its location in the input image. 

The authors of \cite{ma2018docunet} used a straightforward implementation of deducing the backward map by inverting their UV forward map prediction. 
This inversion is done by `placing' the pixels of the forward map in the rectified image based on their values, and performing interpolation over the resulting non-regular grid\footnote{For more details see the Matlab code in \cite{ma2018docunet}}. 
This inversion is very sensitive to noise in the forward map, e.g., if two neighboring pixels swap their predictions.
The same task was addressed in~\cite{li2019docrect} using a parallel iterative method that isn't applicable for an end-to-end differential solution.

We rely on the DenseNet~\cite{huang2017densely} based model provided by~\cite{das2019dewarpnet} for the backward mapper in our solution. The work of \cite{das2019dewarpnet} introduced a learned model for this warp and trained it in a manner independent of the input texture. We rely on their work for transforming our 3D coordinate maps into backward maps. The use of a differentiable backward mapper allows for end-to-end training of the model using a combined objective, optimizing the 3D estimation and backward mapping networks jointly using a combined objective, as discussed in sub-section~\ref{method:optimization}. 

\subsection{Angle Supervision} 
\label{method:angle}

While the 3D estimation network learns the global document structure, we wish the network will also be aware of the local angular deformation that each point of the document has undergone during the warping.
Angular deformation estimation complements the 3D regression used in \cite{das2019dewarpnet} because it is more sensitive to small deformations that might warp parts of words. 
To calculate this value, we warp a local Cartesian system from the source to the target image. 
We use angular deformation estimation in two places in our framework.

\subsubsection{Angle from backward map.}
The first place we estimate the angle is the backward map, where in each pixel we create two infinitesimal vectors $\varepsilon_x$ and $\varepsilon_y$, respectively directed at the $x$ and $y$ directions. 
We then measure the \emph{rotation} that $\varepsilon_x$ and $\varepsilon_y$ undergo due to the warping process, and denote the resulting angles as $\theta_x$ and $\theta_y$.
This process is illustrated in Figure \ref{fig:angle}.
These angles capture the \emph{rotation} and \emph{shear} parts of a local affine transform, without the \emph{translation} and \emph{scale} counterparts that are captured by the coordinate regression. 

\input{figures/angle/angle.tex}

\subsubsection{Auxiliary Angle Prediction.}
In addition to deriving the angles from the backward map, we predict them directly from the 3D estimator network as two auxiliary prediction maps.
These are learned in parallel to the 3D coordinates prediction, as shown in Figure~\ref{fig:arch}, to better guide the training, and are not used during test time. 
Specifically, each of the two angles $\theta_x$ and $\theta_y$ is derived from its own pair of channels, followed by a Cartesian-to-polar conversion (see supplementary material for details).
This conversion yields, in addition to angles $\theta_x$ and $\theta_y$, corresponding magnitude values denoted $\rho_x$ and $\rho_y$.
We use the magnitude values as `angle-confidence' to penalize the angle loss proportionally.
This is beneficial since predictions that have small magnitude are more sensitive to small perturbations.

\subsubsection{Angle Estimation Loss.}
\label{method:loss_angle}
We employ a per-pixel angle penalty on the two aforementioned predictions: one as derived from the backward map, and the other as an auxiliary prediction.
The per-pixel prediction provided by the 3D estimation network is masked by a binary text segmentation map. 
Often, the strongest deformations appear around the borders, and far from content.
Masking content-less areas allows the loss to target areas of interest, and avoid bias towards the highly deformed boundaries.
Our loss minimizes the smallest angle (modulo $2\pi$) between each of the predicted angles $\{\theta_x, \theta_y\}$ and their ground truth counterparts $\{\hat\theta_x, \hat\theta_y\}$. 
The per-pixel loss for angles is therefore:
\begin{eqnarray}
\label{eq:angle_bm}
L_{angle}(\pmb{\theta}, \pmb{\hat\theta}, \pmb{\hat\rho}) = 
     \sum_{i \in \{x, y\}} \hat\rho_i \odot
     ( \|\theta_i - \hat{\theta}_i\| - \pi ) \mod 2\pi,
\end{eqnarray}
where $\odot$ denotes the Hadamard product.
In the backward map angles the loss is without the confidence values $\hat\rho$, that are set to $\pmb{1}$, because the angles are derived from the backward map and are not predicted as an auxiliary.

\subsection{Curvature Estimation} \label{method:crease}
A key observation we utilize in this work is that the surface of a crumpled document behaves in manner similar to a 2D piecewise-planar surface.
Each interface between two approximately-planar sections introduces a section of higher curvature, and higher local distortion.
We wish to give the network a supervision signal that indicates the presence of such high curvature.

Intuitively, the more crumpled the paper, the more creases or discontinuities the warp function exhibits. 
The \emph{curvature map} highlights non-planar areas of the paper, where 3D and angle regression might be less accurate. 
A point in the middle of a plane would have zero curvature, while a point at the tip of a needle would have the maximal curvature value.
To generate this signal, we utilize the 3D mesh used to generate each document image. 

Formally, for a paper mesh $\mathcal{M}$ we calculate a curvature map $H(\mathcal{M})$ using the Laplace-Beltrami operator, as defined for meshes in~\cite{Sorkine2005LaplacianMP}. The mean curvature per mesh vertex $\mathbf{v}_i \in \mathbb{R}^{3}$ is obtained by:
\begin{equation}
H(\mathcal{M})_{i} = ||\sum_{j \in N_{i}} {(\mathbf{v}_i - \mathbf{v}_j)} ||_{2}.
\end{equation}

The maps used for supervision are created by thresholding the curvature, to avoid noise and slight perturbations while emphasizing the actual lines defining the global deformation for the paper. The maps are used as supervision and are predicted as an additional segmentation mask by the 3D estimation network. 
\subsection{Optimization} \label{method:optimization}

The optimization of our model consists of two stages: An initial training stage for the 3D estimation network using the side-tasks, followed by an end-to-end fine tuning stage in which the network is optimized w.r.t. a combined loss term. 

\subsubsection{3D Estimation Model.} Initially, we optimize the 3D estimation model using a loss objective that includes the 3D coordinate estimation loss and the aforementioned auxiliary losses, described in Equation~\eqref{eq:wc_objective}. 
We denote the predicted and ground-truth normalized world coordinates $\mathbf{C}$ and $\mathbf{\hat{C}}$. The first loss term is the $L_1$ loss over coordinate error, similarly to~\cite{das2019dewarpnet}. 
The second term is the angle loss term presented in Equation~\eqref{eq:angle_bm}, masked by the binary text segmentation mask $\mathbf{\hat{D}}$, averaged over all text containing pixels. 
The last term is the curvature estimation $L_2$ loss. The 3D estimation loss is:

\begin{equation}
\label{eq:wc_objective}
    L_{3D} = ||\mathbf{C} - \mathbf{\hat{C}}||_1 + \mathbf{\hat{D}} \odot L_{angle} + ||\mathbf{H} - \mathbf{\hat{H}}||_2.
\end{equation}

\subsubsection{End-to-end Fine Tuning.}
The first stage of our model may either be trained individually, or as part of an end-to-end architecture. When training the model end-to-end, the backward map $B$ is inferred and used for penalizing the predicted 3D coordinates by the final result. 
We penalize the resulting backward map $\hat{B}$ by the $L_1$ loss as was done in \cite{das2019dewarpnet}, and additionally using our angle loss from Equation~\eqref{eq:angle_bm}.
We append these penalty terms to the one in Equation~\eqref{eq:wc_objective}, resulting in the following combined end-to-end loss:
\begin{equation}
\label{eq:e2e_objective}
    L_{combined} = L_{3D} + ||\mathbf{B} - \mathbf{\hat{B}} ||_1 + L_{angle}.
\end{equation}

%% file: figures/arch/arch.tex
\begin{figure*}[t]
    \centering
    \includegraphics[width=0.98\textwidth]{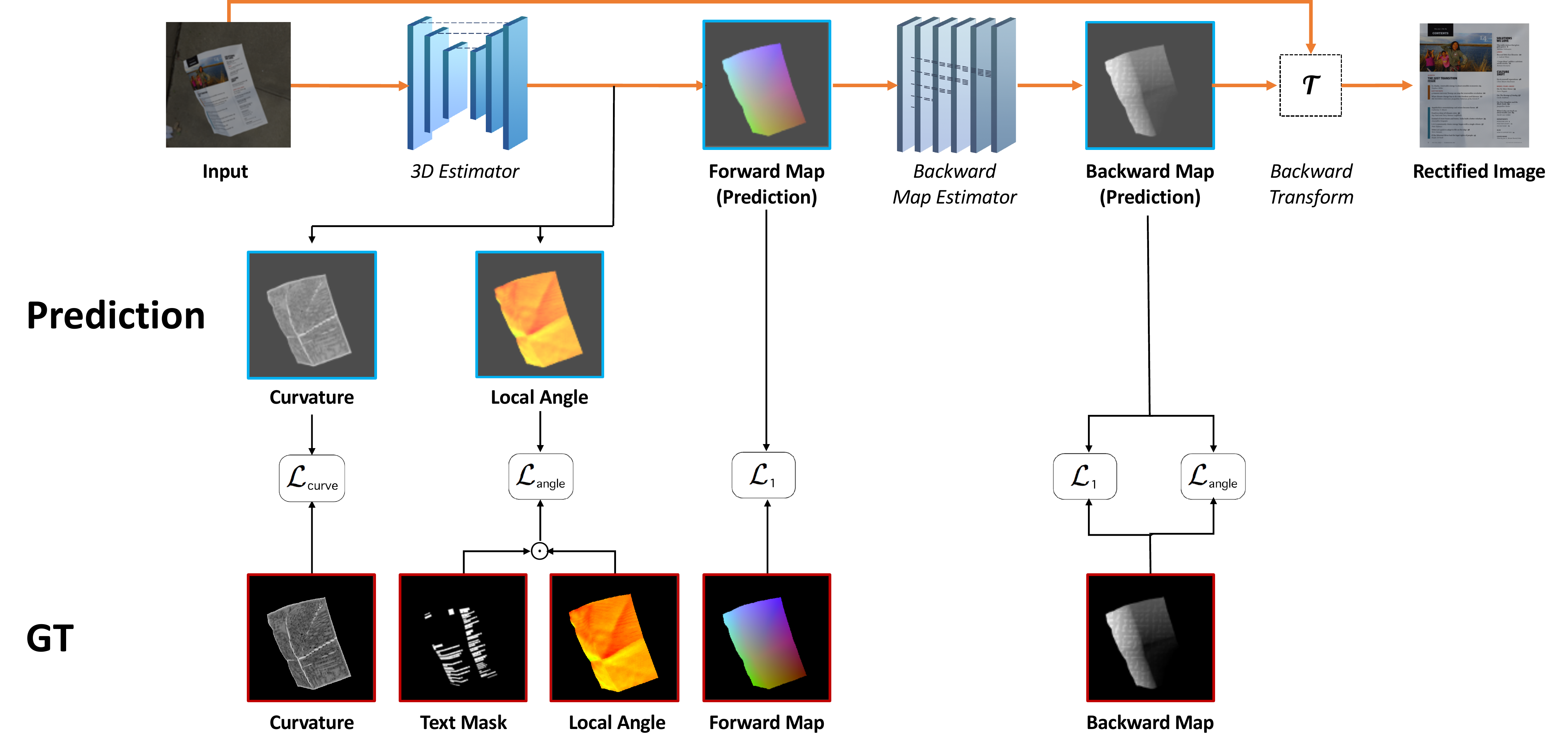}
    \caption{\label{fig:arch}
        {\bf Architecture of CREASE.} 
        The rectification process (in orange arrows) contains two steps: a 3D estimator predicts the 3D coordinates of the document in the image, and a backward map estimator that infers the backward map from the 3D estimation.
        The input image is rectified using the backward map.
        Red and blue frames denote ground truth supervisions and predictions, accordingly. Black arrows denote the training process, and the losses used for optimization. Training is performed first on the 3D estimation model, and is then fine-tuned in an end-to-end fashion.
    }
\end{figure*}

%% file: figures/angle/angle.tex
\begin{figure}[ht]
	\centering
	\begin{overpic}[width=.97\textwidth]{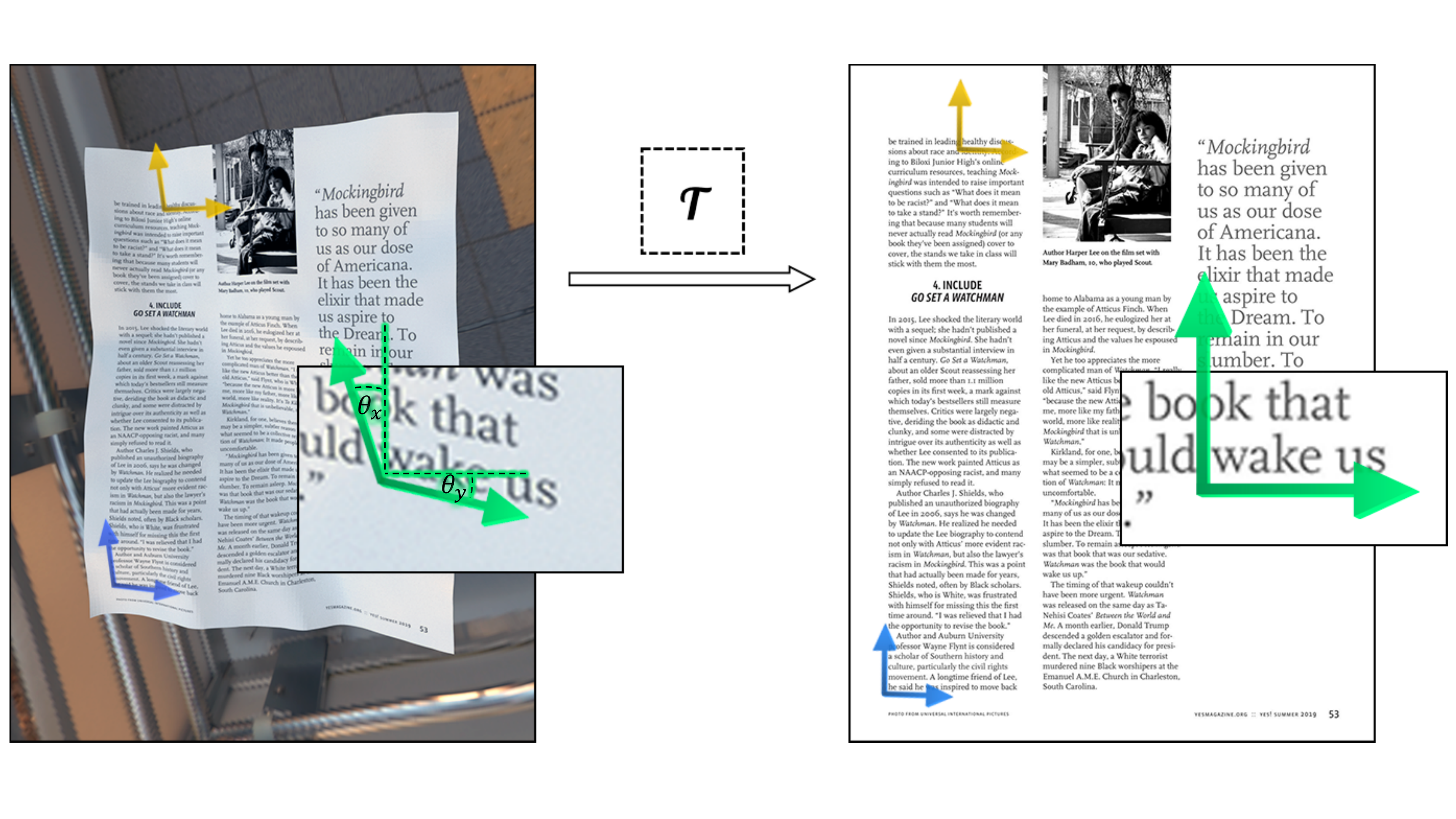}
	\end{overpic}

    \caption{\label{fig:angle}
        {\bf Per-Pixel Angle Calculation. }
        Illustration of the local warp angle calculation in three locations on the document.
        The green point is enlarged to provide better details, showing the resulting angles per axis, denoted by $\theta_x$ and~$\theta_y$.
        This process is used to penalize for angle prediction errors in the the backward map.
        A similar process is done for the forward map, by selecting points from the warped image and transferring them using $\pmb{\tau}$. 
    }    
\end{figure}

%% file: 50_experiments.tex
\section{Experiments}

We evaluate CREASE on a new evaluation set comprised of 50 high resolution synthetic images, as well as on real images from the evaluation set proposed by~\cite{ma2018docunet}. 
The synthetic dataset is generated with both warp and text annotations, useful for OCR based evaluations and for evaluating the individual stages of our model. 
We provide geometric, visual, and OCR based metrics, as well as qualitative evaluations. 
We compare our results to Dewarpnet~\cite{das2019dewarpnet}
trained on our training set, using the code and parameters that were published by the authors. As the method of Li~\etal\cite{li2019docrect} isn't directly applicable to the task at hand, it is not evaluated in this section.
Discussion and comparison to~\cite{li2019docrect} are provided in the supplementary material.

All models were trained on 15,000 high resolution images rendered using an extension of the rendering pipeline provided by~\cite{das2019dewarpnet}. 
Our extensions include the generation of our supervision signals: text, curvature and angles in addition to the 3D coordinates provided by the original rendering pipeline. These added signals come at negligible cost and have no affect in test time. Further details regarding dataset generation are provided in the supplementary material.  

\input{figures/polygon_eval/polygon_eval}
\subsection{Evaluation metrics}

\subsubsection{OCR Based Metric.}
To correctly evaluate any word related metric, we must first obtain a set of aligned word location pairs, i.e., a matching polygon for each ground-truth bounding box in the predicted rectified image domain. 
Given the density and small scale of words in documents, a naive coordinate matching scheme is likely to fail, as a small global shift is to be expected even in the best case scenario.

During evaluation, we rectify an input image twice: using the network's predicted backward map, and using the ground-truth map. 
We then use an OCR engine for extracting words and bounding boxes from the rectified images.

To properly match bounding boxes, we perform the matching stage in the input image domain, visualized in Figure~\ref{fig:polygon_eval}. Each bounding box extracted from a rectified image is warped back and becomes a polygon in the input (warped) image domain.  

We define polygon intersection as our distance metric and match pairs using the Hungarian algorithm~\cite{kuhn1955hungarian}. With the paired prediction and ground-truth word boxes we can evaluate the \emph{Levenshtein distance}~\cite{levenshtein1966binary}, or edit distance, denoted by $E_d$. We first calculate the edit distance for each word in each document, then calculate the average edit distance over all the words in the dataset.

Following \cite{das2019dewarpnet}, we use an off-the-shelf OCR engine (Tesseract $4.0$~\cite{smith2007tesseract}). This engine is quite basic, and does not reflect the advances and robustness of more modern OCR models.
However, the vast majority of recent OCR methods are targeted at scene-text, with the number of proposed text instance detections often limited to 100-200. Thus, they are not suited to handle \emph{dense} document text.
As an alternative, there are a few commercial products designed to handle dense text recognition \cite{gcp_ocr,textract} that are far more advanced than Tesseract. We choose one of them, \cite{textract}, for an additional evaluation.
Results are presented in Tables~\ref{tab:comp} and~\ref{tab:comp_textract}. 

\subsubsection{Geometric and Visual Metrics.}

In addition to an OCR-based evaluation, we use two metrics for evaluating the geometric correctness and visual similarity of our results, \emph{End Point Error (EPE)} and Multi-Scale Structural Similarity (MS-SSIM). 
The EPE metric is used to evaluate the calculated rectification warps and compare them to ground truth. Following~\cite{li2019docrect}, we include evaluation for this metric in our benchmark. 

The MS-SSIM~\cite{wang2004ssim} metric quantifies how visually similar are the output images to the ground truth. 
Given that a small amount of shift is expected and is not considered an error, a naive evaluation using $L_1$ or $L_2$ metrics is not suited for our evaluation.
Therefore, following~\cite{das2019dewarpnet} we use the MS-SSIM metric which focuses on statistical measures rather than per-pixel color accuracy. 
Evaluating statistics rather than per-pixel accuracy also has its limitations, as character level rectification is a fine-grained task and improvements on this scale are not always manifested in this metric.
In fact, SSIM is much more sensitive to small visual deformations in documents containing large amounts of text or sharp edges.
Thus, we only use it to complement our finer-grained, OCR based metrics. For further discussion regarding the SSIM metric, see supplementary. 

\subsection{Implementation Details} \label{experimets:implementation}

Models are trained by first optimizing the 3D estimation network using 3D coordinates, text masks, curvature masks and local angle supervision signals to convergence. Starting from the converged 3D estimation models, we fine-tune our model in an end-to-end manner by using a fixed, pre-trained, differentiable backward mapper. We calculate the $L_1$ and angle losses over the output backward maps and back-propagate the losses to the 3D estimation network. Training is conducted using 15,000 high-resolution images rendered in Blender~\cite{blender} using over 8,000 texture images. 
Further details regarding data generation are provided in the supplementary material.

\input{tables/comparison.tex}
\input{tables/comparison_textract}

\subsection{Comparison to DewarpNet~\cite{das2019dewarpnet}}
The first result we present is a comparison to the prior state-of-the-art trained on our training set, using the Tesseract \cite{smith2007tesseract} engine. 

We show mean and standard deviation values over 5 experiments in Table~\ref{tab:comp}. Our method improves the edit distance metric over the previous method by $4.5\%$ absolute and $20.2\%$ relative. We also see improvements in EPE and SSIM metrics, and a reduction in standard deviation for all three. The use of both angle regression and curvature estimation improves performance and stabilizes the optimization process, reducing the sensitivity to model initialization.

 Next, we evaluate our method using the public online API of \cite{textract}. Results are presented in Table~\ref{tab:comp_textract}. In this case, our model still provides a $5.1\%$ relative improvement. The commercial model \cite{textract} is superior to \cite{smith2007tesseract}, reducing the mean edit distance from $0.178$ to $0.103$, yet CREASE still maintains a significant gap over DewarpNet of $0.6\%$ absolute and $5.1\%$ relative.

\subsection{Evaluation using Real World Images}
Figure~\ref{fig:examples} depicts a qualitative comparison between our rectification method and \cite{das2019dewarpnet} on the real images provided by \cite{ma2018docunet}.
Notice how the text lines rectified using CREASE are better aligned and easier to read than the other method's outputs, especially for text near document edges. Additional examples are included in the supplementary material.

\input{figures/examples/examples}
\input{tables/ablation_angles}

\input{tables/ablation_new}

\subsection{Angle Loss Evaluation}
We show the contribution of the different elements of our angle-based loss presented in Section~\ref{method:angle} for our metrics and for the OCR metric in particular in Table~\ref{tab:ablation_angles}. 
 `Angles' refers to models trained with the angle loss applied to all image pixels, instead of only to those that contain text. 
 `+ Mask' refers to applying the text mask over the loss, i.e., taking the loss only in text-containing pixels, using the mask denoted by $\mathbf{\hat{D}}$ in Equation~\eqref{eq:wc_objective}.
`+ Conf.' represents the use of the angle confidence values (denoted $\rho$ in Equation~\eqref{eq:angle_bm}).
When not used, we set $\rho$ to $1$ for all pixels.
We report results averaged over 5 experiments each, as well as the standard deviation.
For this experiment, the curvature estimation term was omitted. Our contributions show a consistent improvement over the vanilla 3D estimation network and, in addition, a much more stable training framework with consistent results over multiple initializations.

\subsection{Ablation Study} 
Table~\ref{tab:ablation_e2e} shows the effect of each component of our method. Models trained using angle and curvature estimation are compared to vanilla models. We compare both models trained end-to-end (denoted $E2E$) and models trained separately. As seen before, the improvement in results is also accompanied by a decrease in standard deviation, especially for models trained using curvature estimation.
 
We evaluate the contribution of end-to-end training of our model using a fixed, differentiable backward mapper and losses derived from its results, i.e., the backward map and angle prediction errors (shown in Table \ref{tab:ablation_e2e}). 
The top three rows refer to models that were not trained in an end-to-end fashion, while the three rows below (starting with 'E2E') refer to models trained  end-to-end. 'Angles' and 'Curvature' denote the use of each of our two added auxiliary predictions.

The dual usage of the angle loss, in both the 3D estimation model and the end-to-end training, as well as the curvature estimation, result in much more readable rectification and a more stable training scheme than the previous state-of-the-art.

%% file: figures/polygon_eval/polygon_eval.tex
\begin{figure}[t]
\begin{center}

\includegraphics[width=0.32\linewidth]{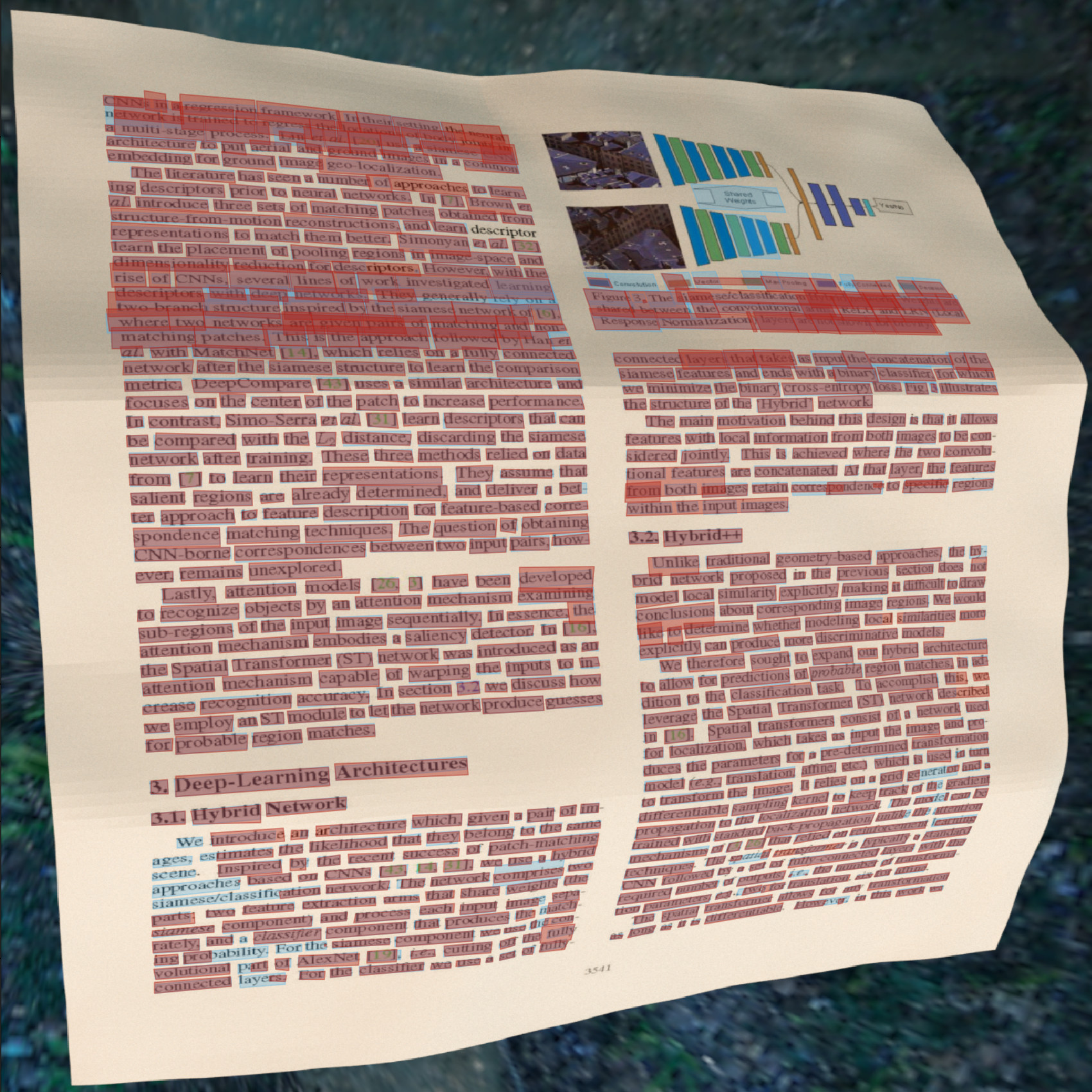}  
\includegraphics[width=0.32\linewidth]{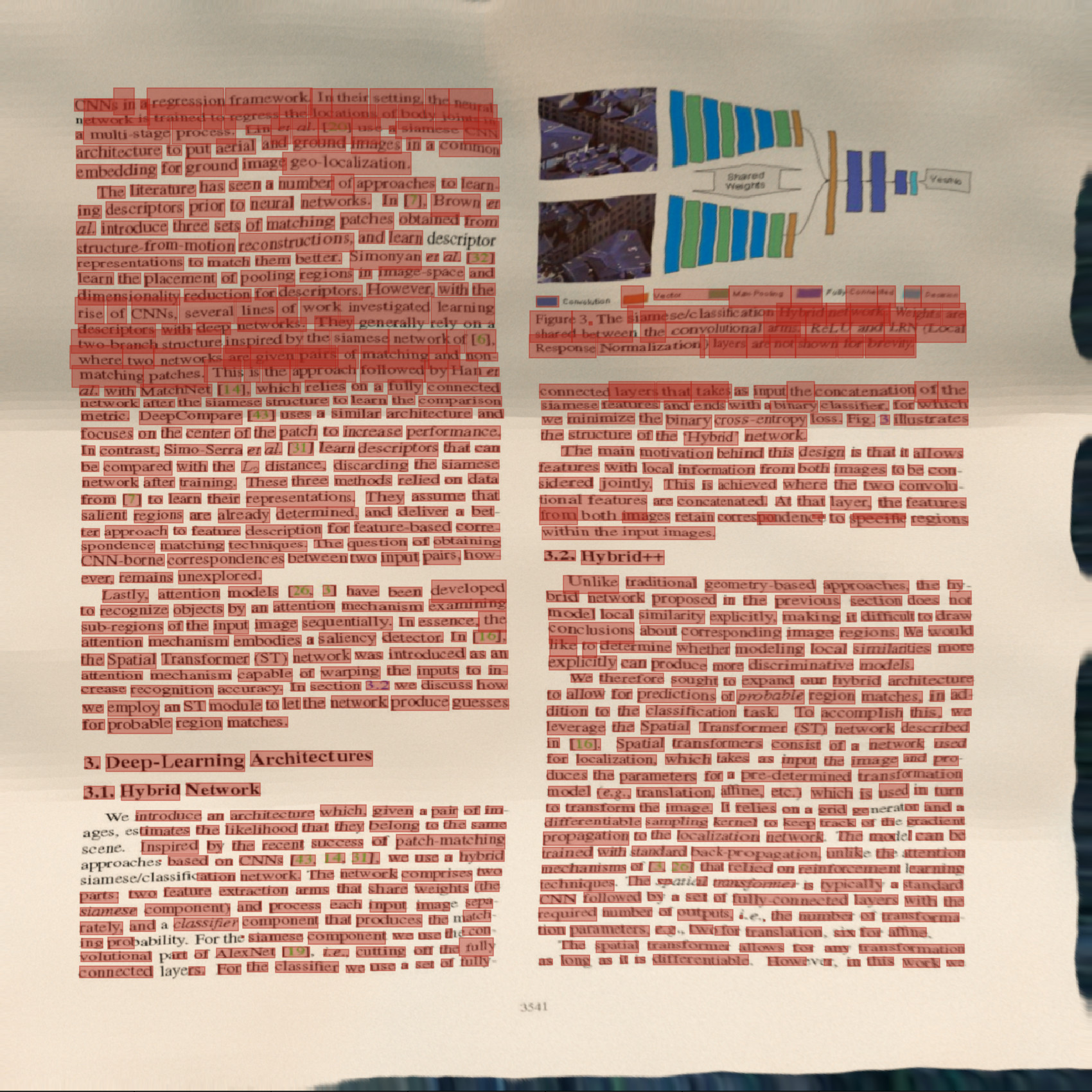}  
\includegraphics[width=0.32\linewidth]{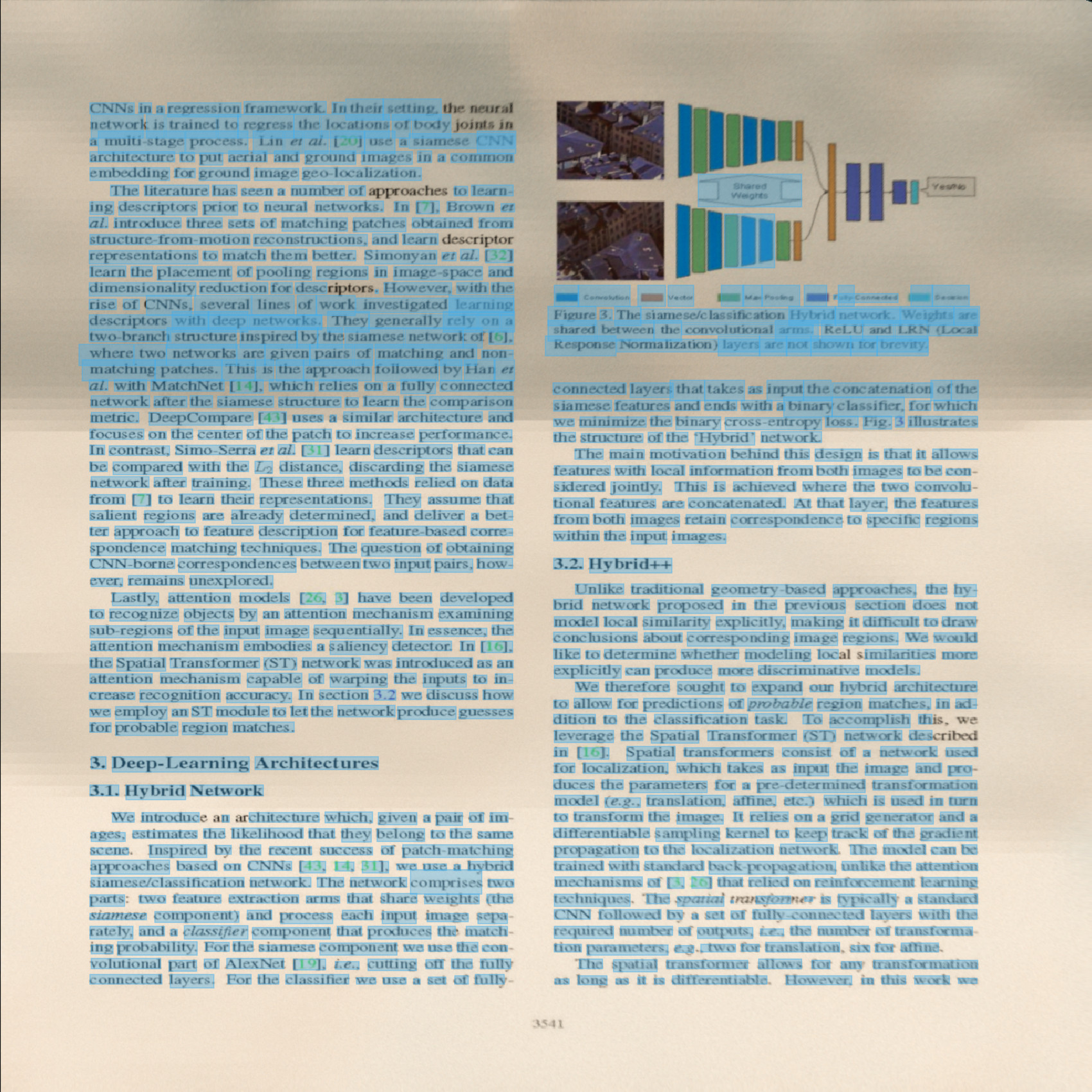}  
\end{center}
\caption{\label{fig:polygon_eval}
    {{\bf OCR Polygon matching. }
    Frames overlayed with OCR bounding boxes.
    Left: Input image overlayed with the warped polygons from images rectified using a predicted transformation (center) and the ground-truth transformation (right).
    Purple areas denote a correct match, blue and red areas were detected only in the ground-truth and prediction rectified images, respectively. 
}
}
\end{figure}

%% file: tables/comparison.tex
\begin{table}[t]
	\setlength{\tabcolsep}{6pt}
	\centering
	\caption{Benchmark Comparison Using Tesseract OCR~\cite{smith2007tesseract}.
	For $E_d$ and EPE, lower is better, while or SSIM, higher is better.}
	\label{tab:comp}
	\begin{tabular}{lccc}
		\hline
		& $\downarrow E_d$ & $\downarrow$ EPE & $\uparrow$ SSIM \\ 
		\hline
		\hline
		DewarpNet \cite{das2019dewarpnet} 
		& 0.223 $\pm$ 0.014 & 0.051 $\pm$ 0.001  & 0.403 $\pm$ 0.004 \\
		
		\hline
		
		Ours  & 0.178 $\pm$ 0.003 & 0.043 $\pm$ 0.002 & 0.411 $\pm$ 0.002 \\
		{\footnotesize Improvement} & +$20.2 \%$ & +$14.1 \%$ & +$1.4\%$ \\
		\hline
	\end{tabular}
\end{table}

%% file: tables/comparison_textract.tex
\begin{table}[t]
\setlength{\tabcolsep}{6pt}
\centering
\caption{Benchmark Comparison Using A Commercial OCR Model \cite{textract}}
\label{tab:comp_textract}
	\begin{tabular}{lc}
		\hline
		 & $\downarrow E_d$  \\ 
		\hline
		\hline
		DewarpNet \cite{das2019dewarpnet} 
		& 0.109 $\pm$ 0.005 \\
		\hline
		Ours & 0.103 $\pm$ 0.001 \\
		{\footnotesize Improvement} & +$5.1 \%$ \\
		\hline
	\end{tabular}
\end{table}

%% file: figures/examples/examples.tex
\begin{figure}[h!]
\centering
 \includegraphics[width=0.92\textwidth]{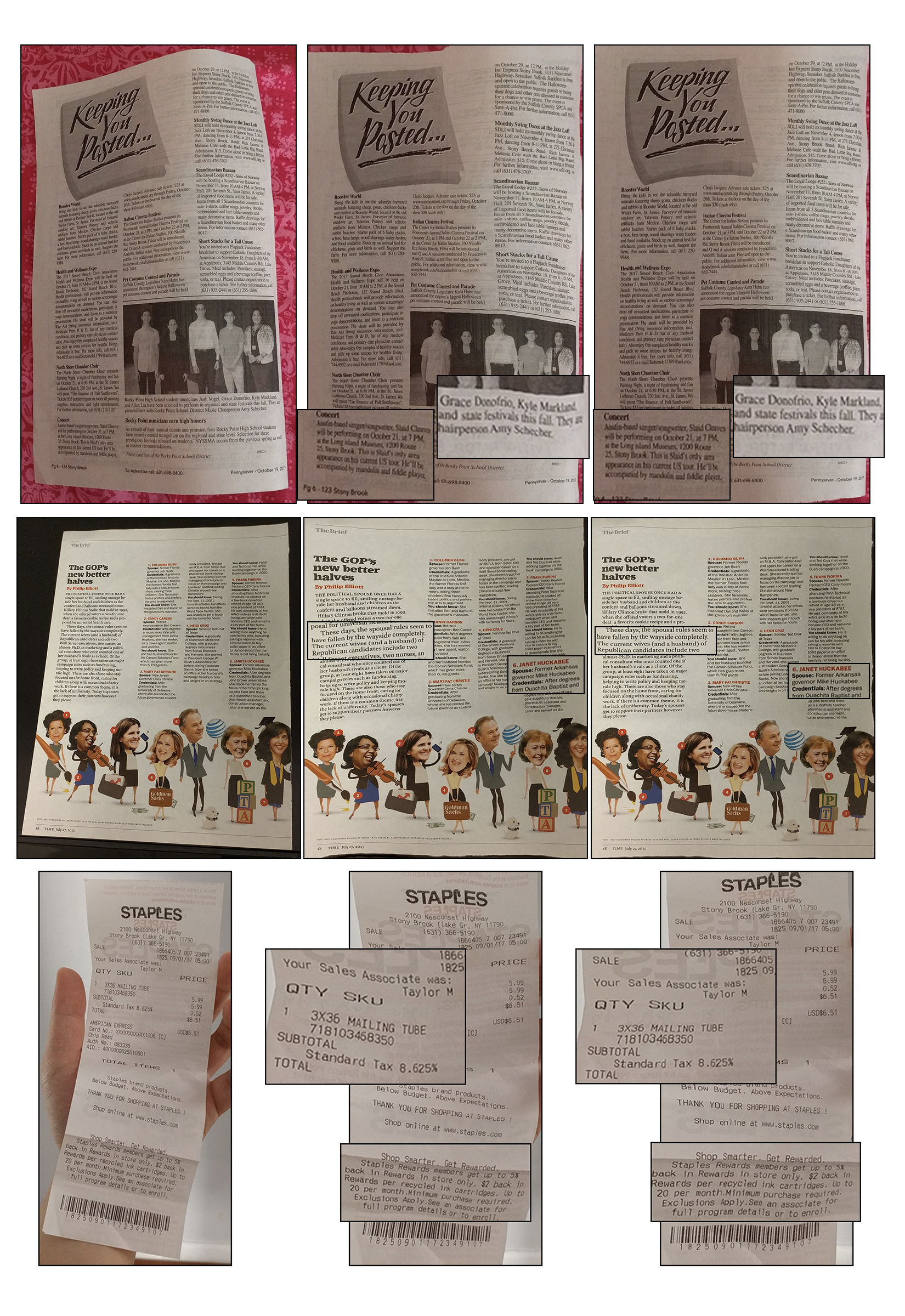} 
\caption{\label{fig:examples}
        {\bf Visual Examples. }
        Results from the real image dataset of~\cite{ma2018docunet}. Left to right: input images, rectification of the input image according to output of~\cite{das2019dewarpnet} trained using our data, rectification of the input image according to our model's output transformation.
}
\end{figure}

%% file: tables/ablation_angles.tex
\begin{table}[t]
\setlength{\tabcolsep}{6pt}
\centering
\caption{Angle Loss Evaluation for the 3D Estimation Model.}
\label{tab:ablation_angles}
	\begin{tabular}{lccc}
		\hline
		Model & $\downarrow E_d$ & $\downarrow$ EPE & $\uparrow$ SSIM  \\
		\hline
		\hline
		Vanilla     & 0.324 $\pm$ 0.169 & 0.066 $\pm$ 0.029 & 0.390 $\pm$ 0.024 \\
		\hline
		Angles      & 0.246 $\pm$ 0.017 & 0.049 $\pm$ 0.002 & 0.403 $\pm$ 0.005 \\ 
		+ Mask      & 0.244 $\pm$ 0.014 & 0.052 $\pm$ 0.001 & 0.398 $\pm$ 0.005 \\
		+ Conf.     & 0.216	$\pm$ 0.017 & 0.049 $\pm$ 0.001 & 0.400 $\pm$ 0.005 \\
		\hline
	\end{tabular}
\end{table}

%% file: tables/ablation_new.tex
\begin{table}[t!]
\setlength{\tabcolsep}{6pt}
\centering
\caption{Ablation Study.}
\label{tab:ablation_e2e}
	\begin{tabular}{lccc}
		\hline
        & $\downarrow E_d$ & $\downarrow$ EPE & $\uparrow$ SSIM  \\ 
		\hline
		\hline
		Vanilla         & 0.324 $\pm$ 0.169 & 0.066 $\pm$ 0.029 & 0.390 $\pm$ 0.024 \\
		Angles          & 0.216	$\pm$ 0.017 & 0.049 $\pm$ 0.001 & 0.400 $\pm$ 0.005 \\
		Angles + Curvature & 0.187 $\pm$ 0.005 & 0.043 $\pm$ 0.001 & 0.409 $\pm$ 0.007 \\
		
		\hline
		E2E                   & 0.223 $\pm$ 0.014 & 0.051 $\pm$ 0.001 & 0.403 $\pm$ 0.004 \\
		E2E + Angles          & 0.204 $\pm$ 0.015 & 0.051 $\pm$ 0.002 & 0.402 $\pm$ 0.005 \\
		E2E + Angles + Curvature & 0.178 $\pm$ 0.003 & 0.043 $\pm$ 0.002 & 0.411 $\pm$ 0.002 \\
		\hline
	\end{tabular}
\end{table}

%% file: 60_conclusion.tex
\section{Conclusion}

We presented CREASE, a content aware document rectification method which optimizes a per-pixel angle regression loss, a curvature estimation loss and a 3D coordinate estimation loss for providing image rectification maps. 

Our method rectifies folded and creased documents using hints found in both local and global scale properties of the document, and provides a significant improvement in OCR performance, geometry and visual similarity based metrics. 
In our proposed two stage model, the first stage is used for predicting 3D structure, angles and curvature, while the second stage predicts the backward map.
We utilize a pixel-level angle regression loss that is shown to be a beneficial side-task in both the 3D estimation and the end-to-end training.
Furthermore, our 3D estimation model learns the angle side-task specifically on the words in the document, thus optimizing for readability in the rectified image, while the curvature estimation side-task complements the angle regression by mapping its discontinuities.

Extensive testing and comparisons show our method's superior performance over diverse inputs, using both real and synthetic evaluation data. We show an increase in OCR performance, geometry and similarity metrics that is consistent over all experiments and on a variety of documents.

%% file: supp/100_supp_intro.tex
\begin{center}
    \LARGE{\bf Supplementary Material}
\end{center}

The supplementary material provides a comparison to the method proposed by Li \etal\cite{li2019docrect} (Section~\ref{supp:docrect}), details regarding the dataset used for training (Section~\ref{supp:data_gen}), and more information regarding the Cartesian to Polar conversion done during the calculation of our loss (Section~\ref{supp:car2pol}). In addition, we present a side-task providing word segmentation masks (Section~\ref{supp:text_seg}) and discuss the usefulness of the MS-SSIM metric (Section~\ref{supp:ssim}) and provide more visual results from our method. 

%% file: supp/110_supp_docrect.tex
\section{Comparison to Li~\etal\cite{li2019docrect}} \label{supp:docrect}

A recent related work by Li~\etal\cite{li2019docrect} presented a method for document rectification focused on uneven background illumination and gently folded documents. The work took a patch-based approach for inferring local flow fields followed by a graph-cut model for stitching the patches back to the complete flow. The results obtained using this method both using their publicly available model and by models re-trained using our data are availabe in Table.~\ref{supp_tab:comp_docrect}.

While this method shares several similarities with work presented and compared to in this paper, it is not suitable for the kind of deformations found in our dataset and handled both by our model and by DewarpNet~\cite{das2019dewarpnet}. Faced with complex deformations that render non-planar patches, it often resorts to inconsistent stitching patterns, as seen in Fig.~\ref{supp_fig:docrect_fail}. 

Due to the method's incomparable results we have decided to include it separately, and evaluate both the authors' publicly available pre-trained model\footnote{\url{https://github.com/xiaoyu258/DocProj}}, and the averaged results of 5 models trained using our dataset.

\input{supp/tables/comparison_docrect}
\input{supp/figures/docrect/docrect_samples}

%% file: supp/tables/comparison_docrect.tex
\begin{table}[h]
\centering
\caption{Benchmark Comparison to~\cite{li2019docrect} on Synthetic Data. Mean results and standard deviation over the test set. (\textdagger)~denotes the author's pre-trained model. (*)~denotes models retrained using our training set. For SSIM, higher is better.
For $E_d$ and EPE, lower is better.}
\label{supp_tab:comp_docrect}
\addtolength{\tabcolsep}{5pt}

\begin{tabular}{lccc}
\hline

          & ED & SSIM & EPE \\ 
\hline
\hline

Li~\etal\cite{li2019docrect}$^\dagger$ &   0.683   &  0.281 & 0.205   \\
Li~\etal\cite{li2019docrect}$^*$       & 0.652 $\pm$ 0.027 & 0.263 $\pm$ 0.003 & 0.184 $\pm$ 0.003 \\

\hline

CREASE
  & 0.178 $\pm$ 0.003 & 0.411 $\pm$ 0.002 & 0.043 $\pm$ 0.002 \\
  
\hline
\end{tabular}
\end{table}

%% file: supp/figures/docrect/docrect_samples.tex
\begin{figure}
\begin{center}
\begin{tabular}{cccc}
\includegraphics[width=0.24\linewidth]{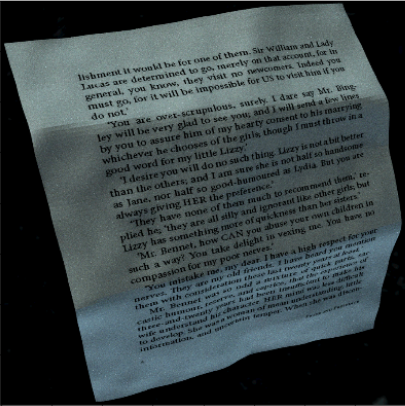} &
\includegraphics[width=0.24\linewidth]{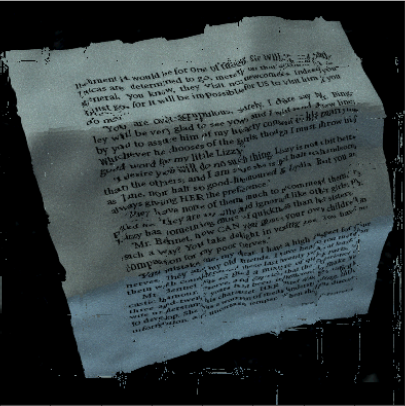} &
\includegraphics[width=0.24\linewidth]{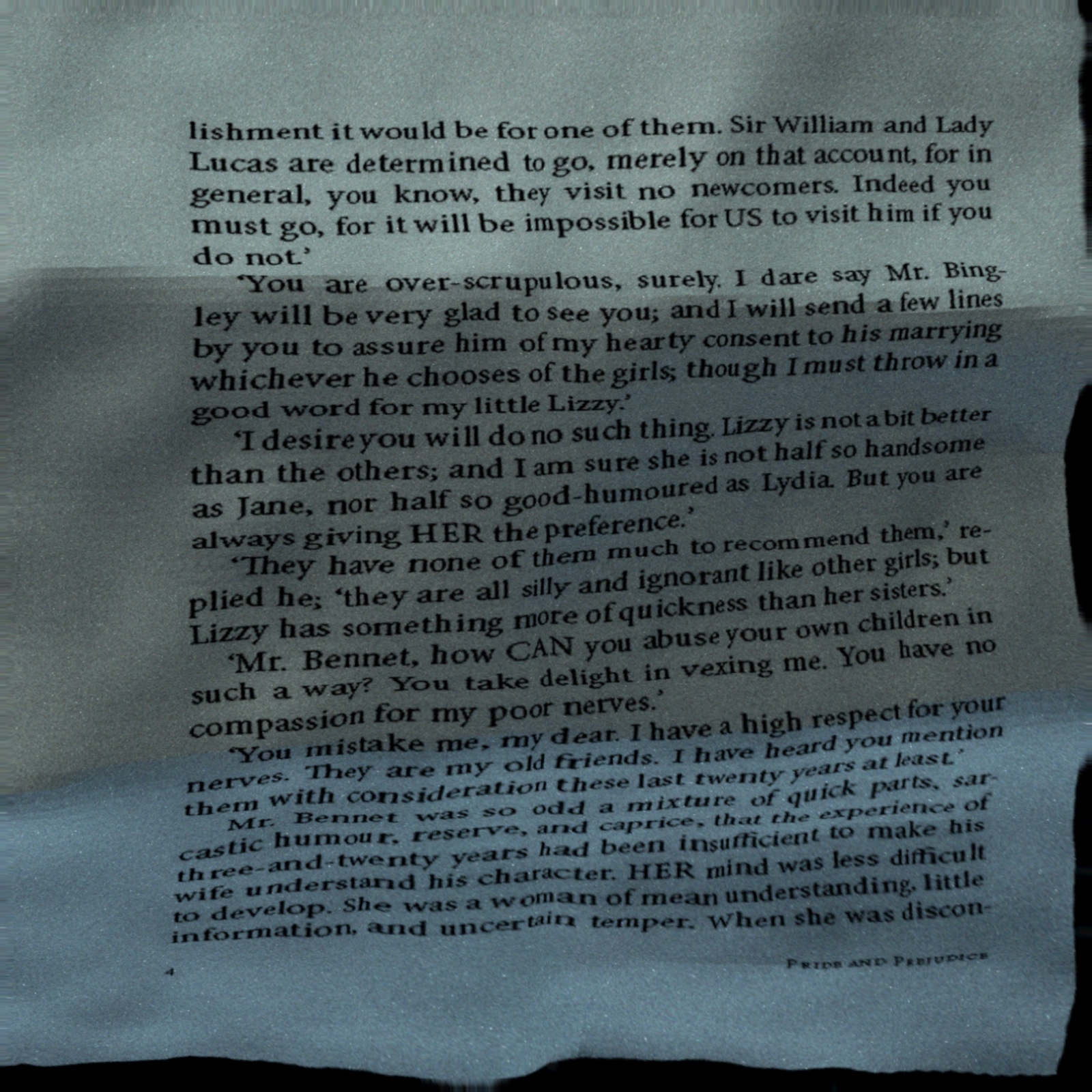} & 
\includegraphics[width=0.24\linewidth]{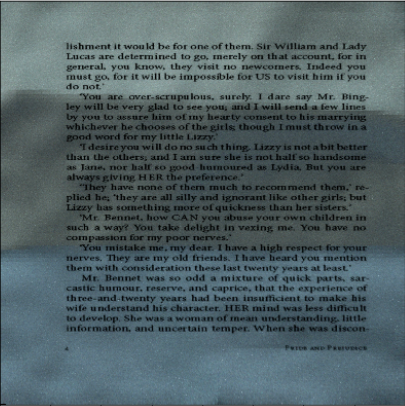}  \\

Input & Li~\etal\cite{li2019docrect} & Ours & Ground-Truth \\
\vspace{-0.30cm}
\end{tabular}

\end{center}
\caption{\label{supp_fig:docrect_fail}
    {\small \textbf{Results From Li~\etal\cite{li2019docrect}.} 
    Left to right: Input image, Results from~\cite{li2019docrect}, Our results, and the result of rectifying the input image using the ground-truth backward map. Notice uneven boundaries and visible stitches in the patch based method.
}
\vspace{-0.5cm}}
\end{figure}

%% file: supp/120_supp_dataset.tex
\section{Training Dataset} \label{supp:data_gen}

\subsubsection{Generation of our dataset.} The data was generated in Blender~\cite{blender} using 10,000 document images and 8000 meshes. The doucment images were extracted from PDFs collected from open-access magazines, books, academic papers, in multiple formats (one, two and three columns, advertisements with a single text blob, etc.) that include diverse images, figures and text.
The 8,000 meshes are those used in~\cite{das2019dewarpnet}, and were kindly provided by the authors. In addition to the warped images, 3D world coordinates, and UV maps provided by the renderer, we extracted the content meta-data from each PDF document, including text and word bounding boxes, and used them to create the flattened binary text masks. In the following step, each flat mask was warped using the generated backward map. Unlike previous works that rendered relatively low resolution images, here images and annotations were rendered in a $1600 \times 1600$ pixel resolution, useful for fine grained OCR evaluations and closer to real world scanning resolutions.

 \subsubsection{Doc3D} The authors of~\cite{das2019dewarpnet} presented the Doc3D dataset, that was also generated in Blender in a similar manner to ours. In Doc3D, however, a few limitations prohibited us from using it during our training and evaluation protocols: 
 (i) At the time of writing, the former dataset is no longer publicly available, except for the meshes used for generation;
 (ii) The dataset was generated in a $448 \times 448$ resolution, significantly below the required threshold for OCR and even unreadable by people for commonly used font sizes.
 We thus retrained the models of~\cite{das2019dewarpnet} using our dataset and the training parameters from the publicly available implementation\footnote{\url{https://github.com/cvlab-stonybrook/DewarpNet}}.

%% file: supp/130_supp_cart2polar.tex
\section{Cartesian to Polar Coordinate Conversion} \label{supp:car2pol}

In order to apply our angular deformation estimation based loss, we predict the rotation angle of each of the two axes. 
The use of a separate angle for each axis corresponds to both rotation and shear. 
In other words, during the deformation, both axes rotate differently on a per-pixel level. Axes are rotated individually, and are no longer orthogonal as in a flat surface. The predicted maps account for the magnitudes of the change of each axis, in each direction.  

Specifically, the 3D estimation model predicts 4 auxiliary channels, a pair for each axis, which we denote $(\phi_{xx}, \phi_{xy}, \phi_{yx}, \phi_{yy})$. The predictions of $\phi_{xy}$ provide the value of shift predicted for the $X$ axis in the $Y$ direction, and so forth. For each axis, we then calculate the angle $\theta_i$, and the magnitude $\rho_i$ for $i \in {x, y}$: 

\label{supp_eq:cart2polar}
\begin{equation}
    \theta_i =  \arctantwo(\phi_{ix}, \phi_{iy}) ,
\end{equation}
\begin{equation}
    \rho_i =  ||(\phi_{ix}, \phi_{iy})||_2 ,
\end{equation}
where `$\arctantwo$' is the four-quadrant variation of the arctangent operator (also referred to as `$\atantwo$'). The calculated values are then used in our loss, as described in Section $3.2$ in our paper.

%% file: supp/135_supp_text_seg.tex
\section{Word Segmentation Output} \label{supp:text_seg}
We train an auxiliary word segmentation channel as part of the 3D estimation model. We show in Table \ref{tab:text} that this channel does not improve our results on the OCR metrics. However, this channel can quite accurately localize words and lines areas, as seen in Figure~\ref{fig:text_seg}. This can be beneficial for the next task in the pipeline, e.g. text localization in the document. 
\input{tables/text_segmentation.tex}
\input{figures/text_seg/text_seg}

%% file: tables/text_segmentation.tex
\begin{table}[h]
\setlength{\tabcolsep}{4pt}
\centering
\caption{Text Segmentation Auxiliary Channel}
\label{tab:text}

{\small 
\begin{tabular}{lccc}
\hline

          & $E_d$ & EPE & SSIM \\ 
\hline
\hline
No Text & 0.178	$\pm$ 0.003 & 0.043 $\pm$ 0.002 & 0.411 $\pm$ 0.002 \\
Text    & 0.182 $\pm$ 0.003 & 0.043 $\pm$ 0.002 & 0.409 $\pm$ 0.004  \\
\hline
\end{tabular}
} 
\end{table}

%% file: figures/text_seg/text_seg.tex
\begin{figure*}[h]
    \centering
    \begin{tabular}{cccc}
            \includegraphics[height=0.18\textwidth]{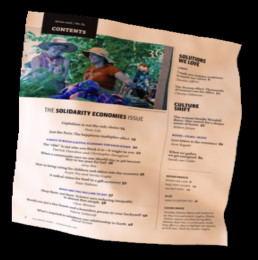} &
            \includegraphics[height=0.18\textwidth]{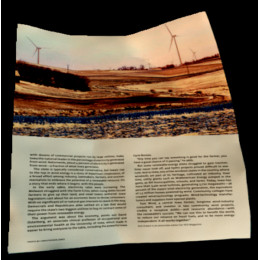} &
            \includegraphics[height=0.18\textwidth]{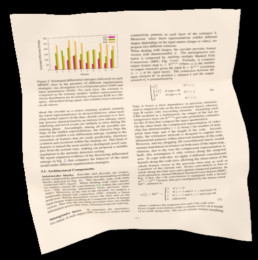} &
            \includegraphics[height=0.18\textwidth]{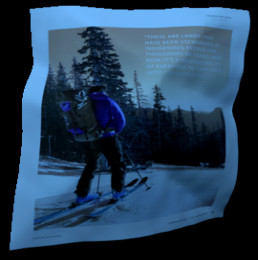} \\ 
            \includegraphics[height=0.18\textwidth]{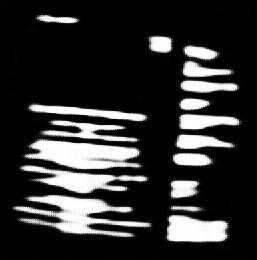} &
            \includegraphics[height=0.18\textwidth]{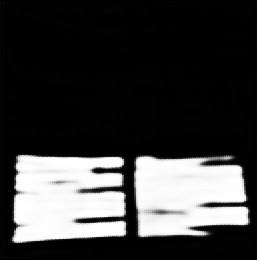} &
            \includegraphics[height=0.18\textwidth]{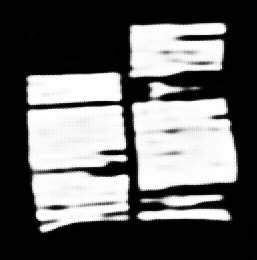} &
            \includegraphics[height=0.18\textwidth]{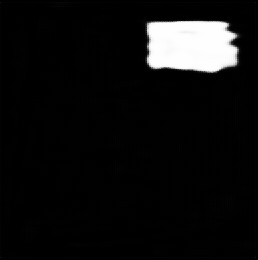} \\ 
            \includegraphics[height=0.18\textwidth]{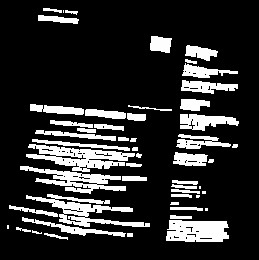} &
            \includegraphics[height=0.18\textwidth]{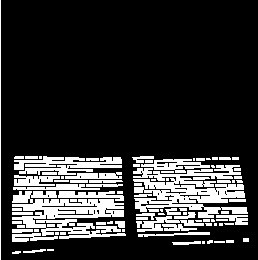} &
            \includegraphics[height=0.18\textwidth]{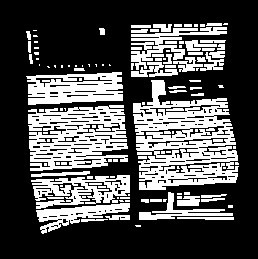} &
            \includegraphics[height=0.18\textwidth]{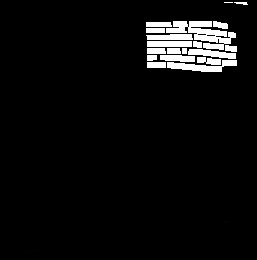} \\ 
    \end{tabular}
    \caption{\label{fig:text_seg}
      A visualization of the text segmentation output provided by our 3D estimation model.  
      Top row shows input images, middle row shows our model's predictions, and ground-truth predictions are presented in the bottom row. Robustness of this output can be seen in the right-most column, where text is properly segmented even in relatively low contrast areas.
    }
\end{figure*}

%% file: supp/140_supp_ssim.tex
\section{The MS-SSIM Metric} \label{supp:ssim}

The MS-SSIM~\cite{wang2004ssim} metric was used in this work and in other works to capture the rectified document's similarity to the ground truth, rectified variant of it. Given a small shift is to be expected even in very accurate predictions of the backward map, common per-pixel comparison metrics such as $L_1$ and $L_2$ are not useful for the task, as they would require an additional non-trivial registration step. 

The MS-SSIM metric was chosen as an alternative for evaluating global similarity, specifically when used in multi-scale and applied over an image pyramid. SSIM is an alternative to $L_1$/$L_2$ in the sense that it is more correlative with human perception. SSIM still, however, suffers for the same need for exact registration the nultiscale MS-SSIM might deal with this issue more gracefully than the original counterpart.

However, the SSIM based metrics present their own disadvantages, and specifically, lack of sensitivity to fine-grained details and a bias towards even textures. As seen in Fig.~\ref{supp_fig:ssim}, a document that wasn't rectified, but contains relatively even and flat textures (including background textures showing)
exhibits a far superior similarity score to a properly rectified document that contains dense text. This comes to show that, while useful and intuitive for many applications, in the specific case of document rectification, the MS-SSIM metric isn't fully suitable for comparing the fine-grained details required for a useful, readable output. 

\input{supp/figures/ssim/ssim_fig}

%% file: supp/figures/ssim/ssim_fig.tex
\begin{figure}
\begin{center}
\begin{tabular}{ccc}
Input & Output & Ground-Truth  \\
\includegraphics[width=0.3\linewidth]{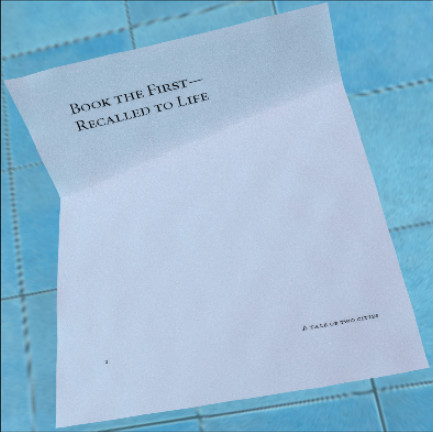} &
\includegraphics[width=0.3\linewidth]{supp/figures/ssim/ssim_high_in.jpg} &
\includegraphics[width=0.3\linewidth]{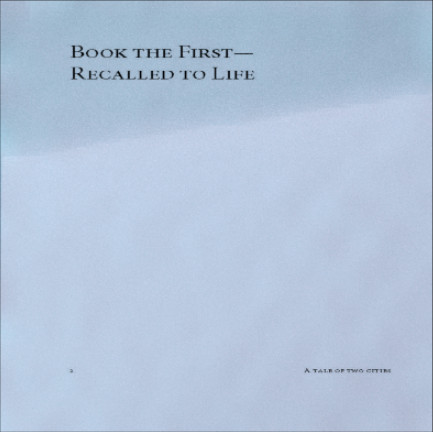}  \\
& MS-SSIM = $0.625$ & \\
& $E_d = 0.750 $ \\ \\

\includegraphics[width=0.3\linewidth]{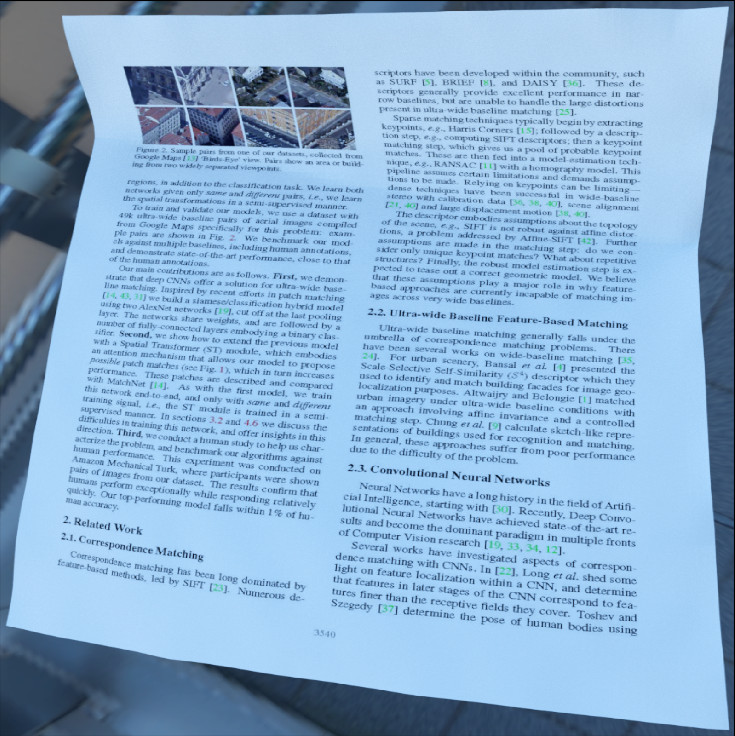} &
\includegraphics[width=0.3\linewidth]{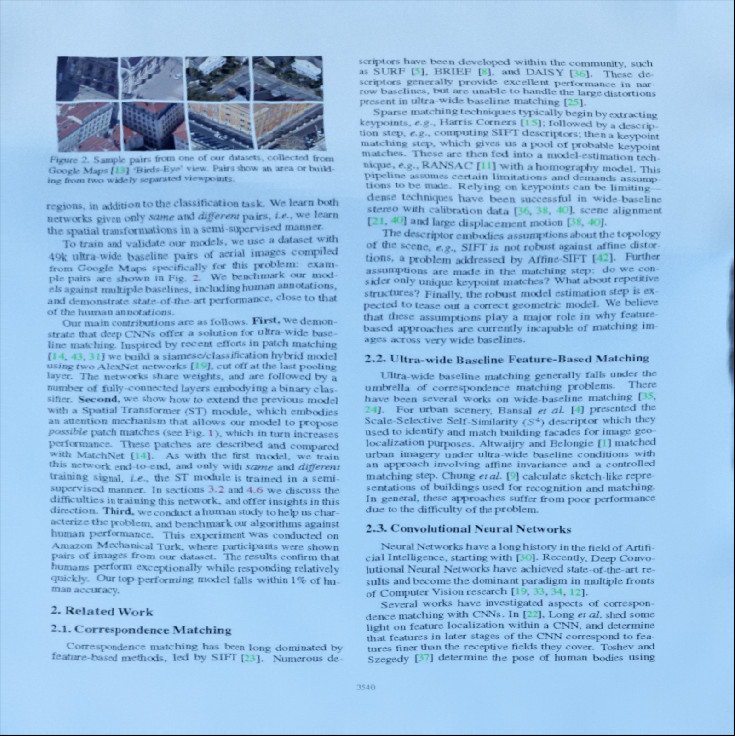} &
\includegraphics[width=0.3\linewidth]{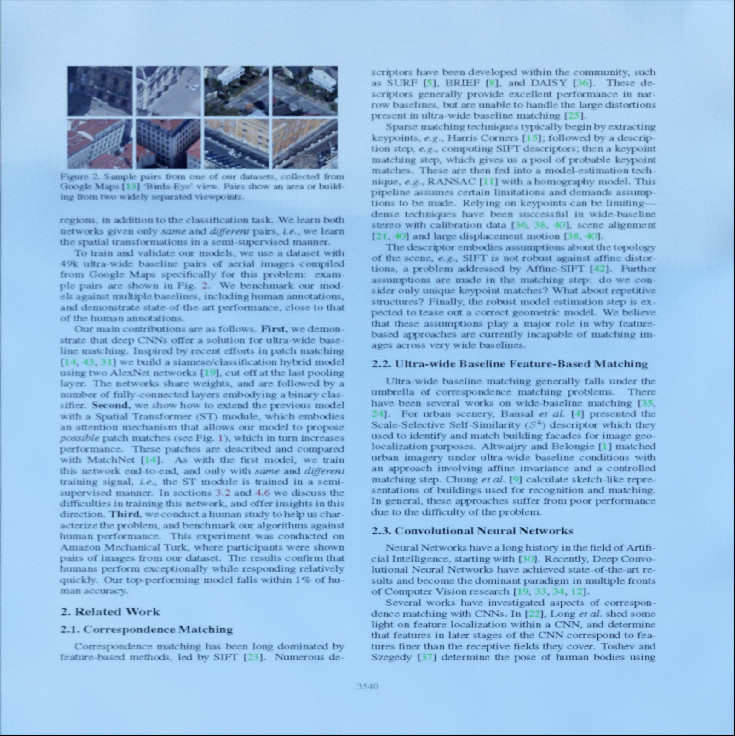}  \\
& MS-SSIM = $0.265$ & \\
& $E_d = 0.042$ \\ \\
\end{tabular}

\end{center}
\caption{\label{supp_fig:ssim}
    {\small \textbf{Bias in MS-SSIM.} 
    \emph{Top row:} we compare the input image directly, with and identity mapping (no rectification), to the ground-truth rectified image.
    The high SSIM score shows the bias of the metric towards even surfaces, even when rotated and when a large portion of background is showing.
    \emph{Bottom row:} Input image is rectified using our method. It is then compared to the ground-truth rectified image. 
    Notice, that even for a successful rectification yielding very low $E_d$, SSIM value is significantly lower.
}
\vspace{-0.2cm}}
\end{figure}

%% file: supp/150_supp_vis_results.tex
\label{supp:additional_results}

\input{supp/figures/supp_samples/supp_samples}

%% file: supp/figures/supp_samples/supp_samples.tex
\begin{figure}
\begin{center}
\begin{tabular}{ccc}
Input & Vanilla & CREASE  \\
\includegraphics[width=0.28\linewidth]{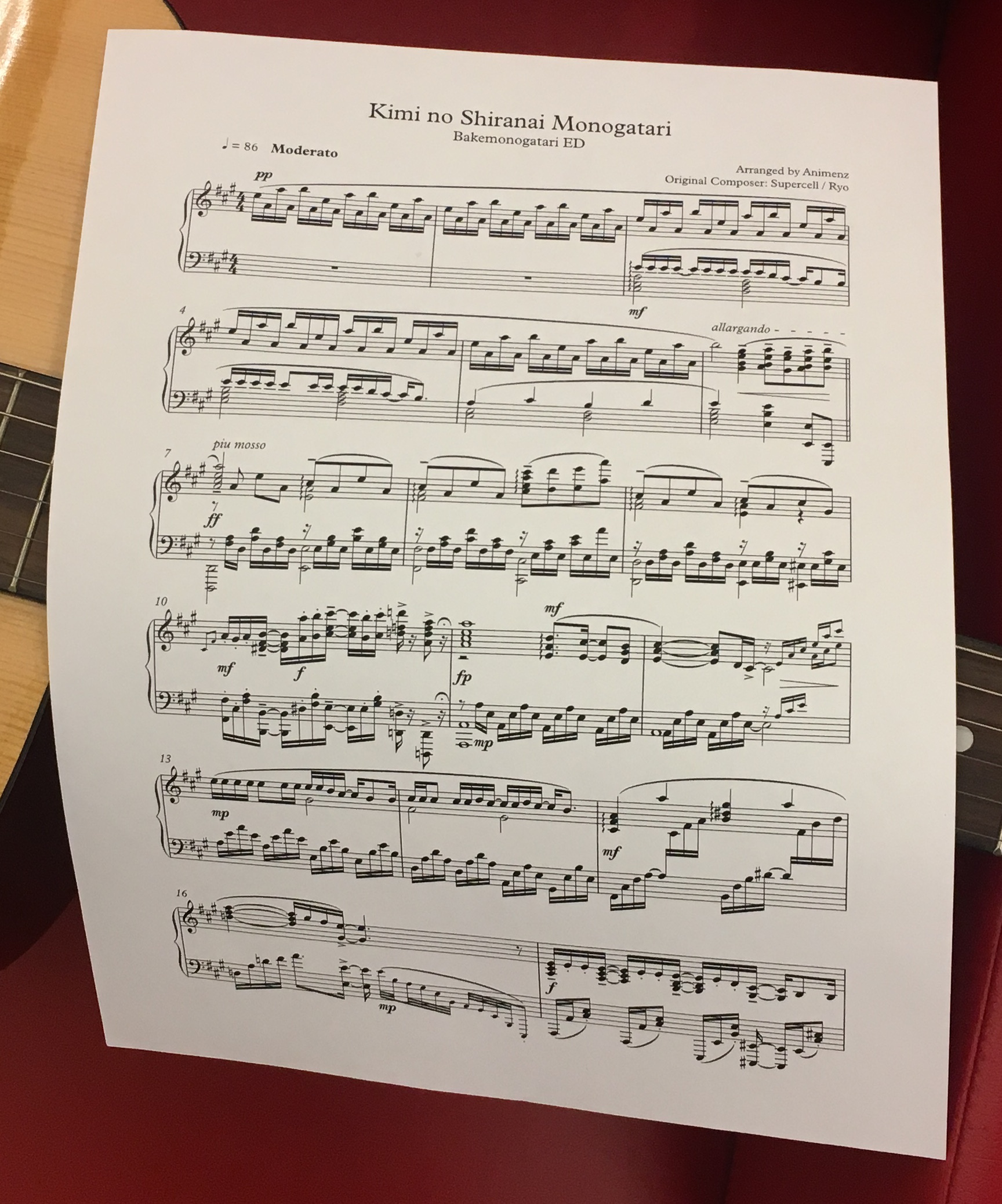} &
\includegraphics[width=0.28\linewidth]{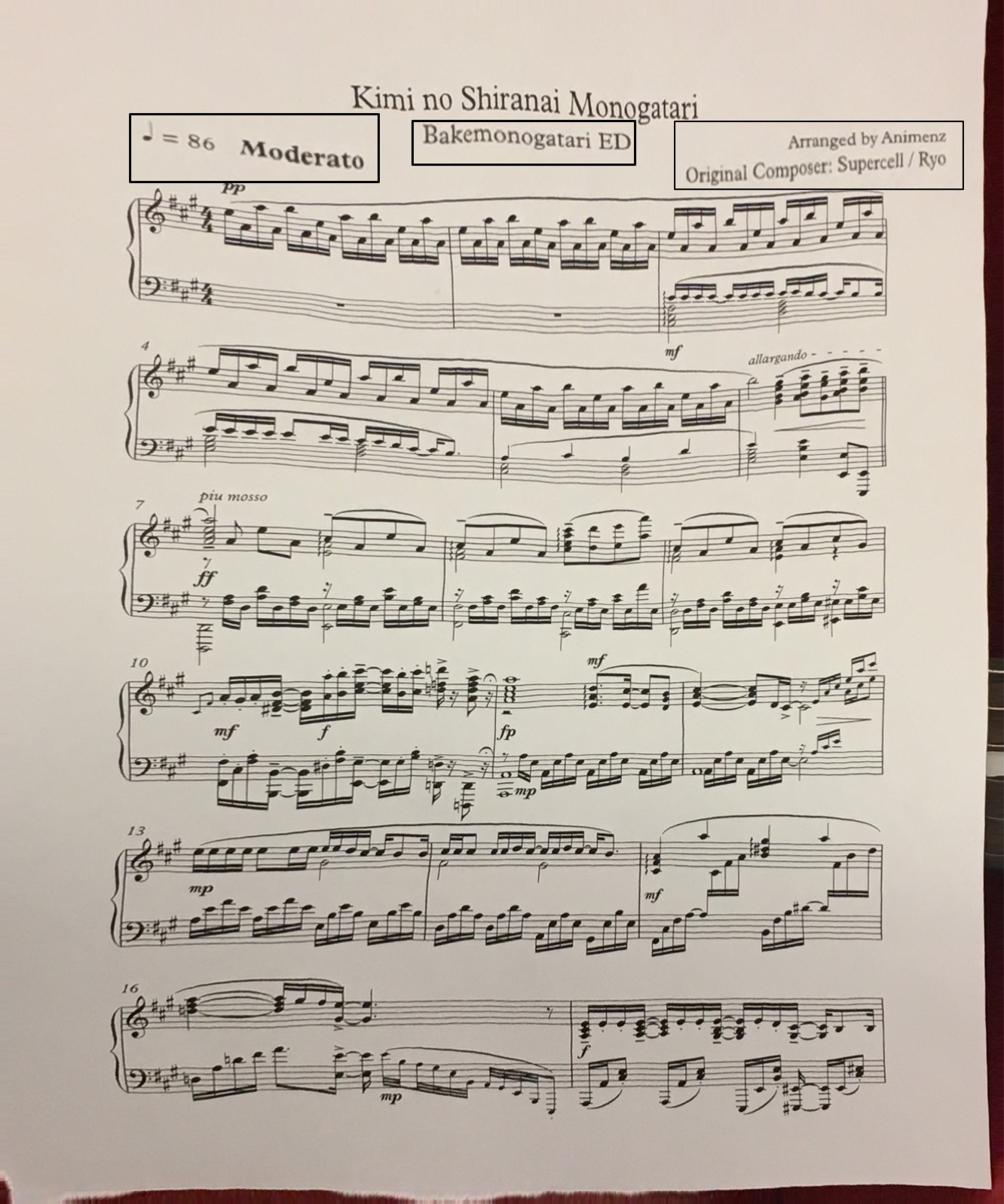} &
\includegraphics[width=0.28\linewidth]{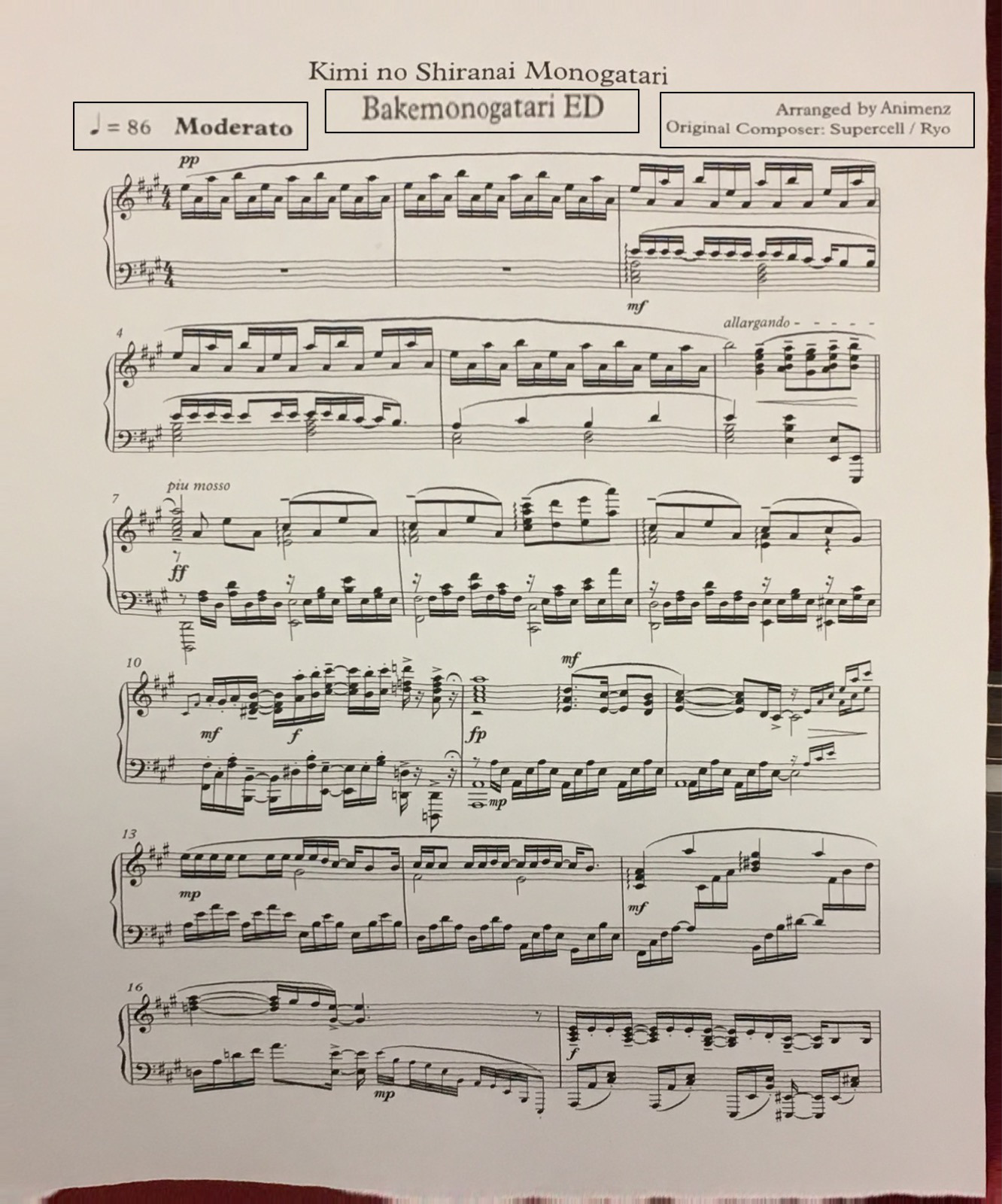} \\

\includegraphics[width=0.28\linewidth]{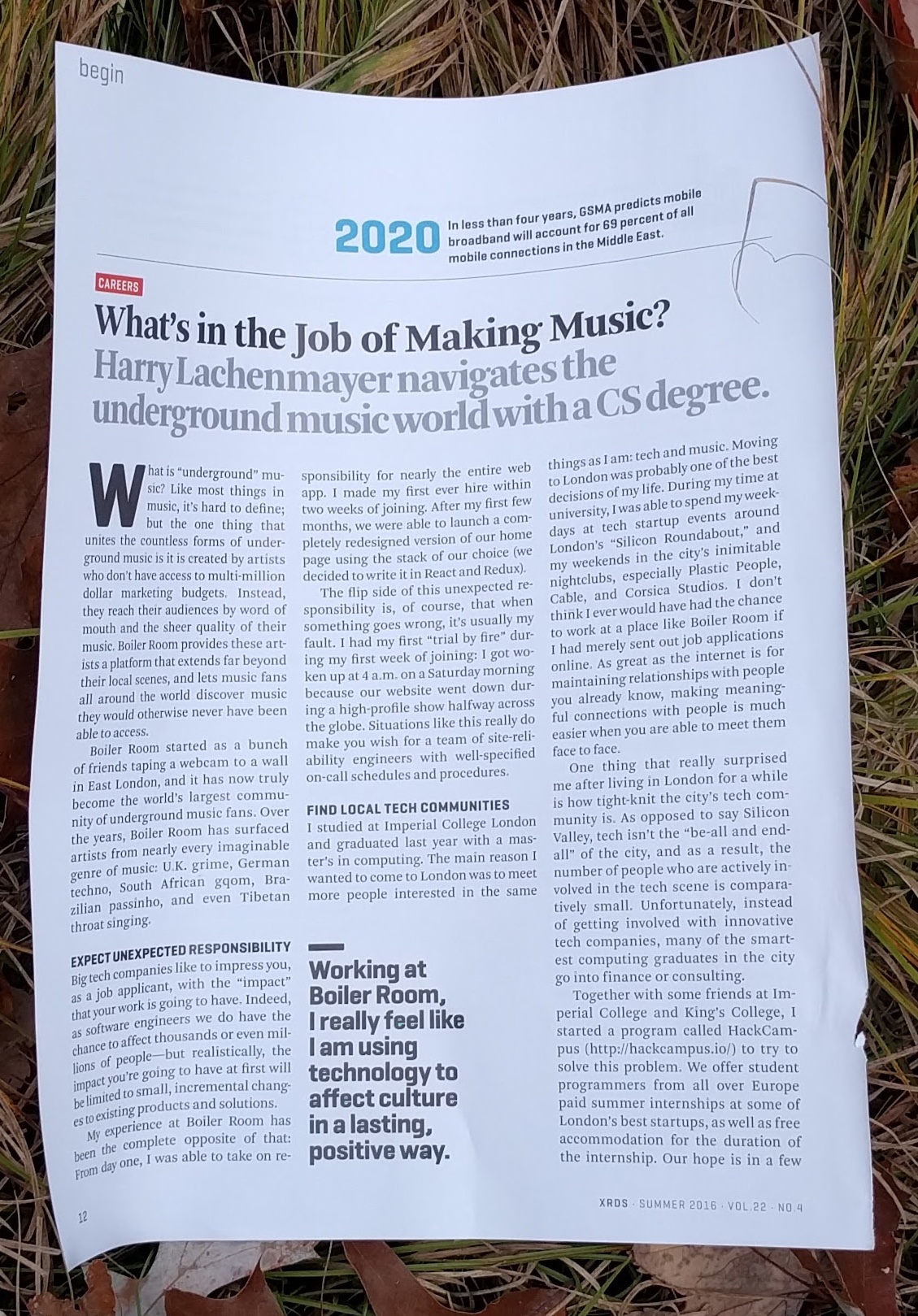} &
\includegraphics[width=0.28\linewidth]{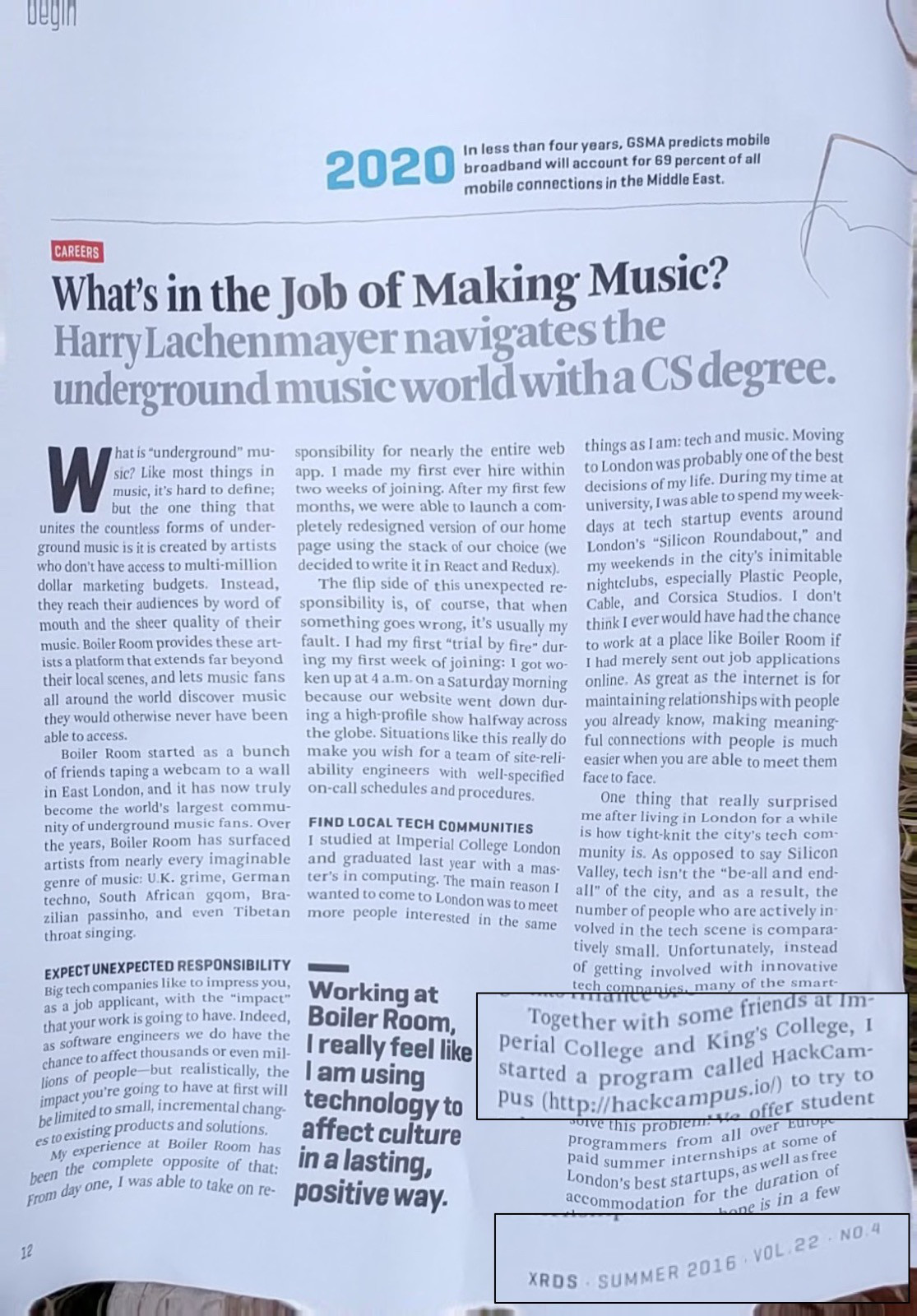} &
\includegraphics[width=0.28\linewidth]{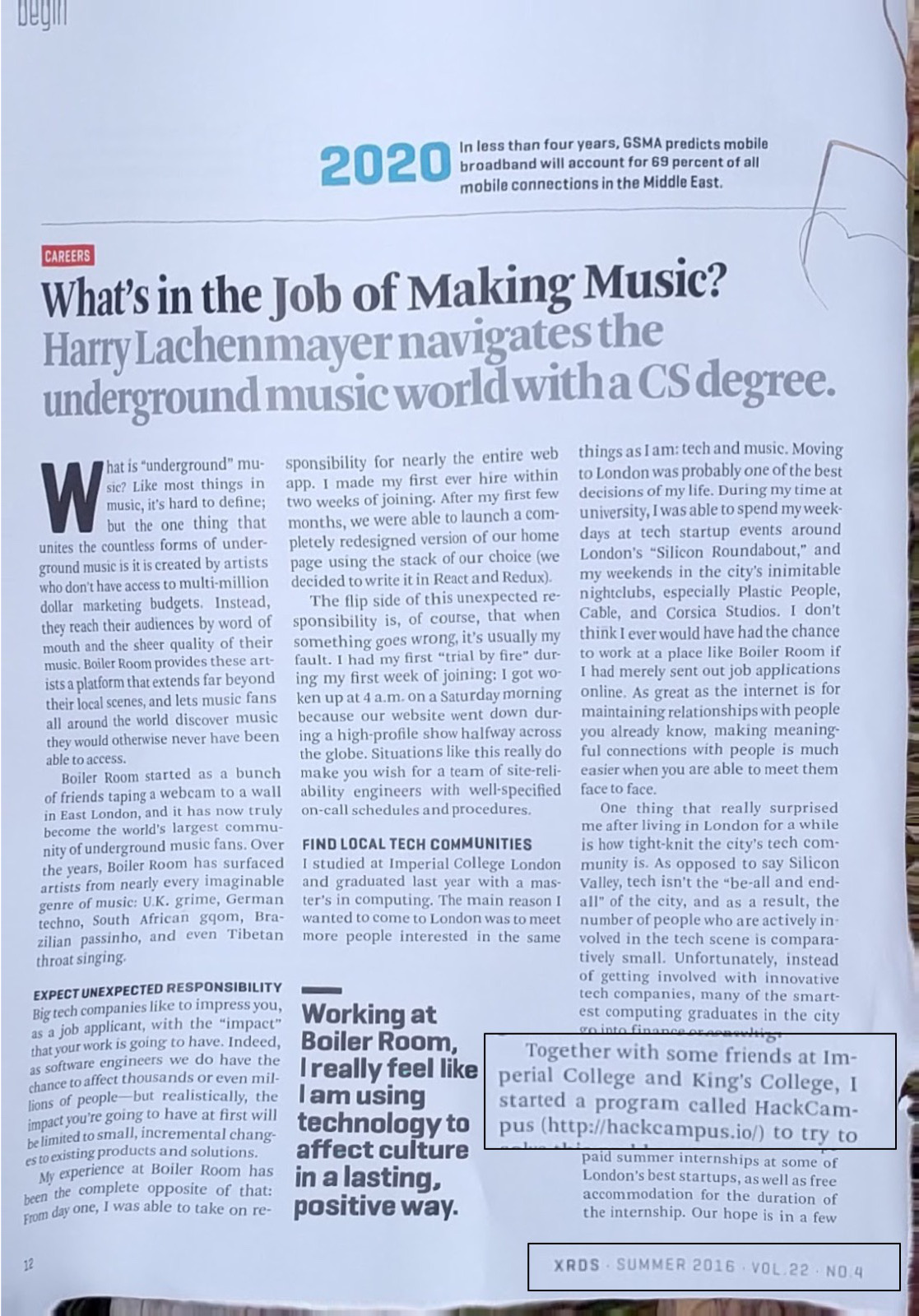} \\

\includegraphics[width=0.28\linewidth]{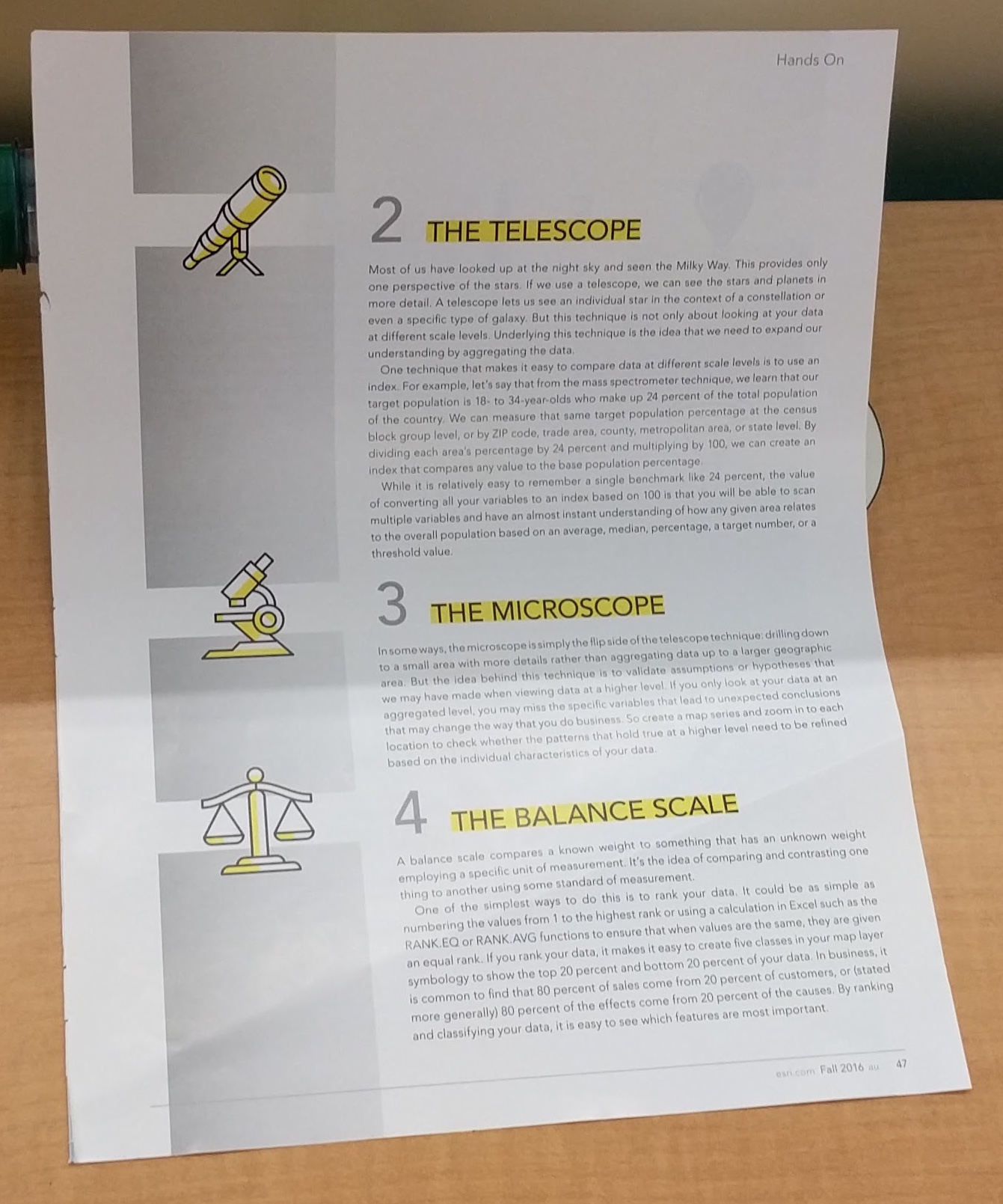} &
\includegraphics[width=0.28\linewidth]{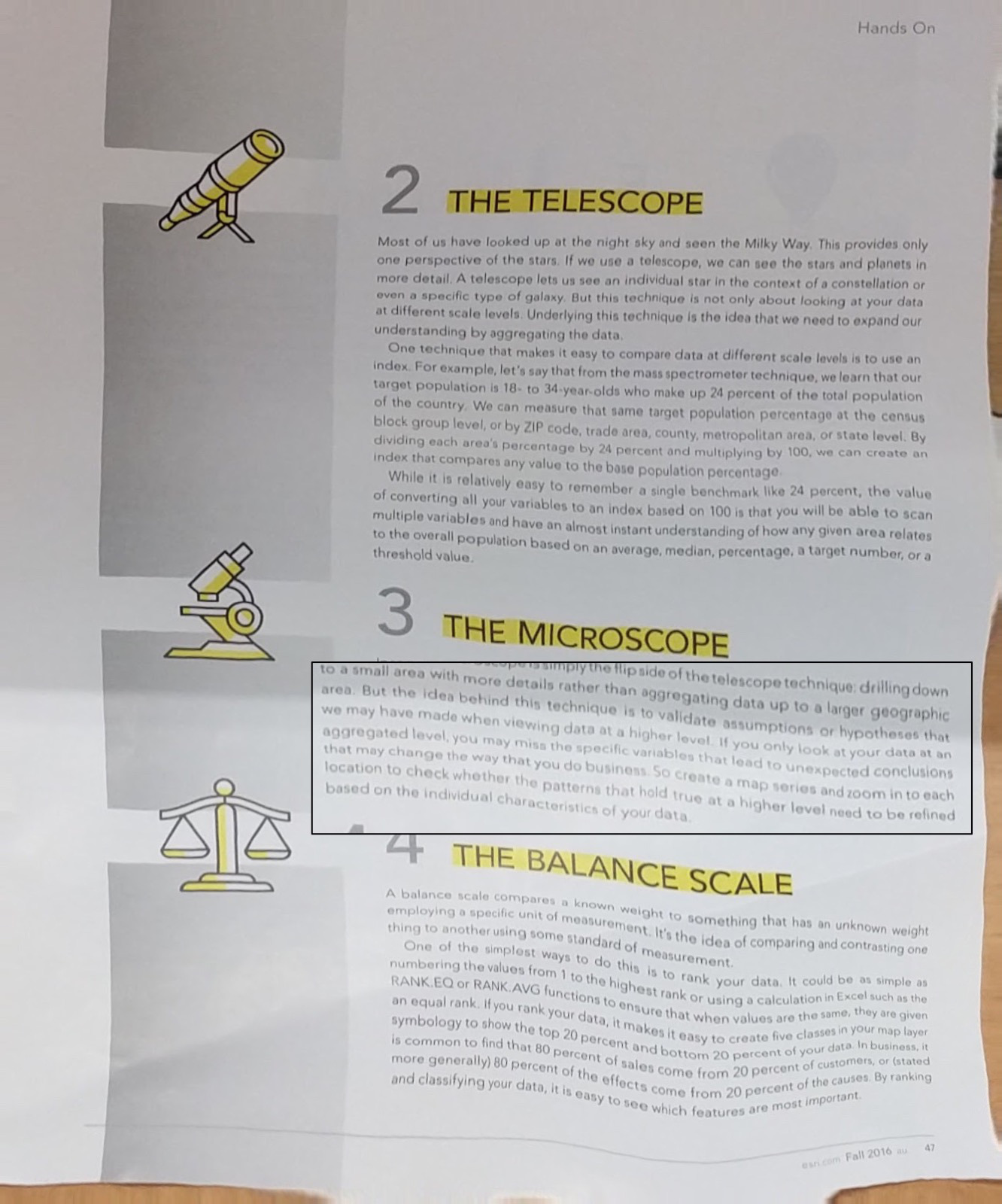} &
\includegraphics[width=0.28\linewidth]{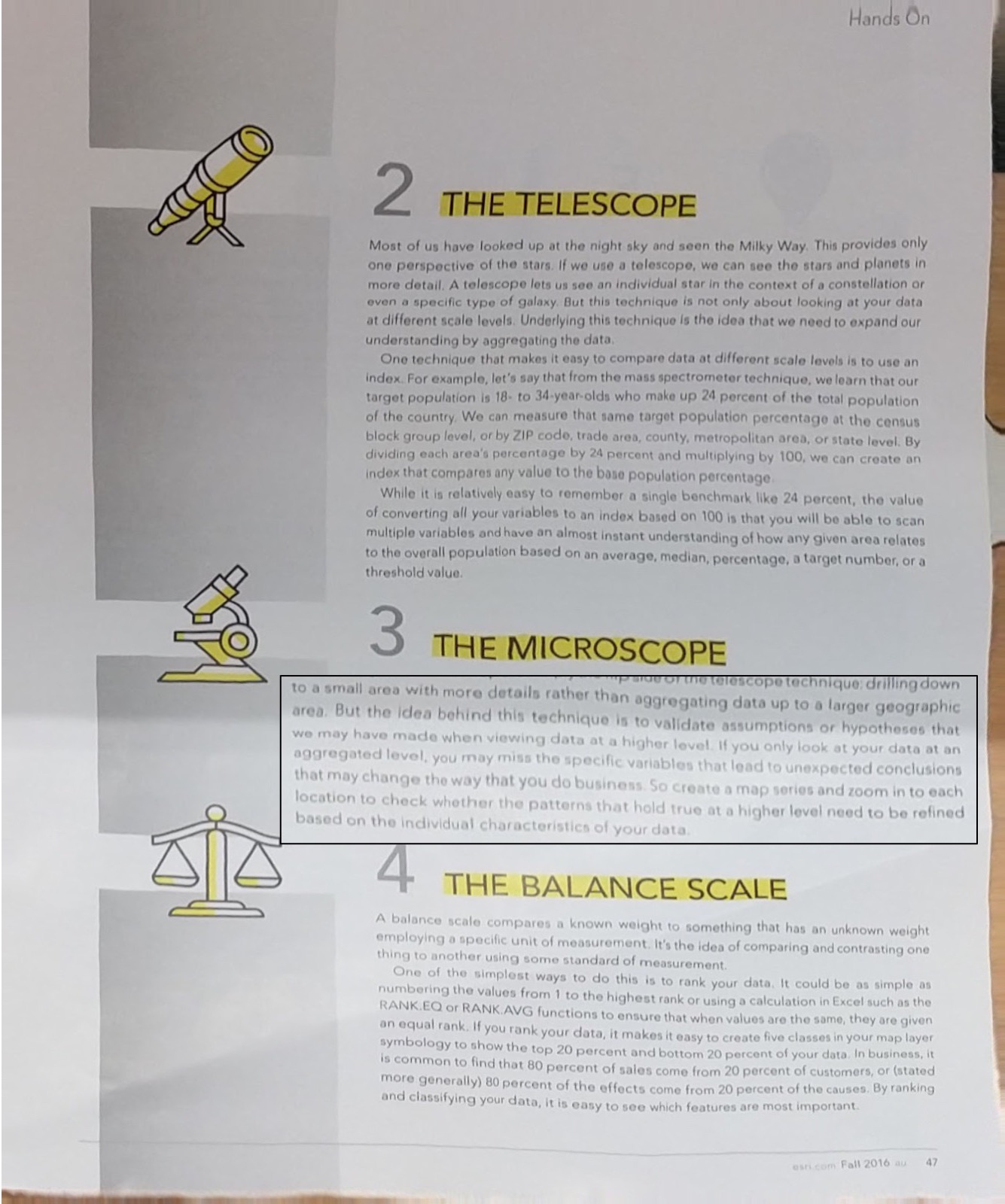} \\

\includegraphics[width=0.28\linewidth]{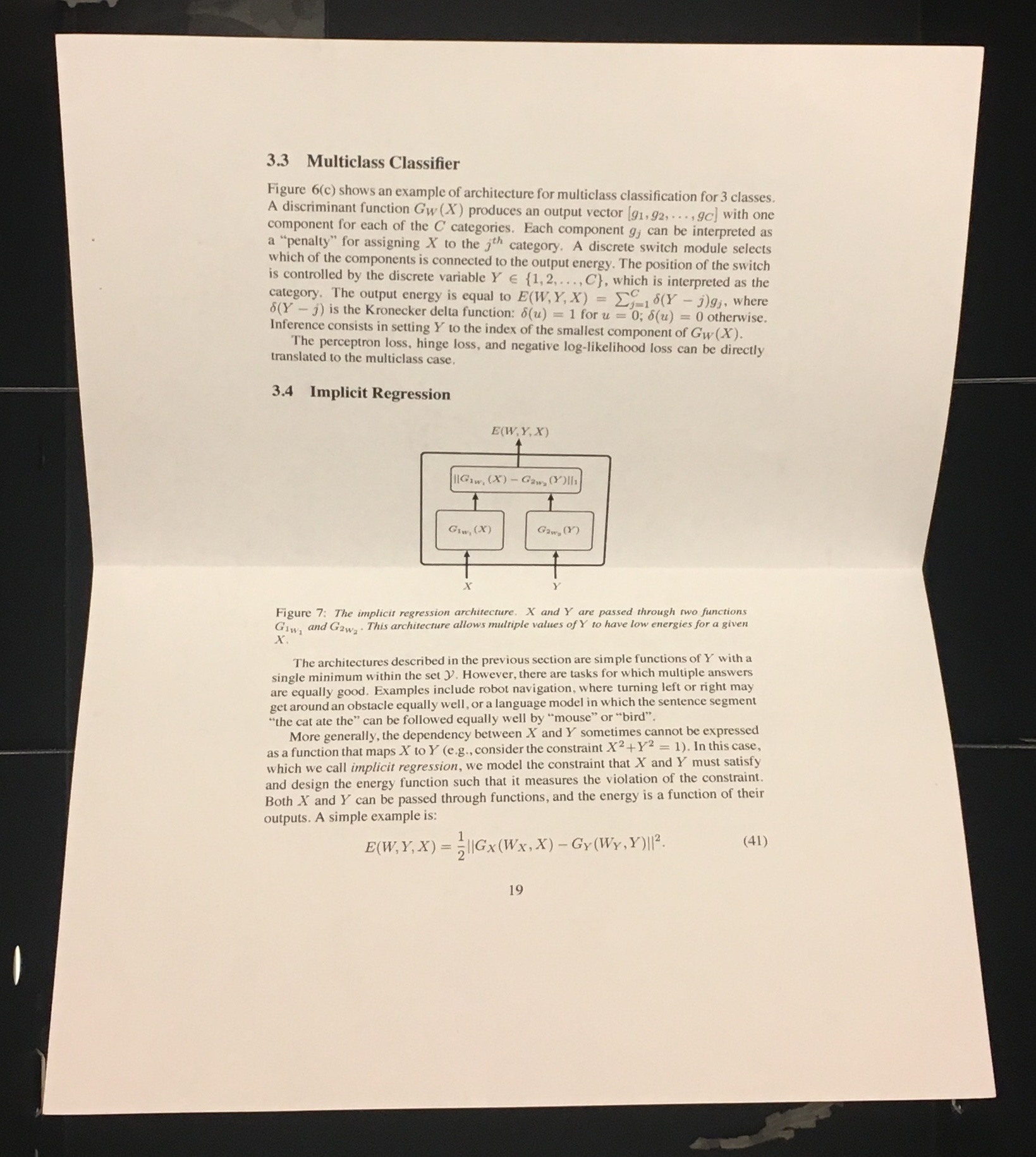} &
\includegraphics[width=0.28\linewidth]{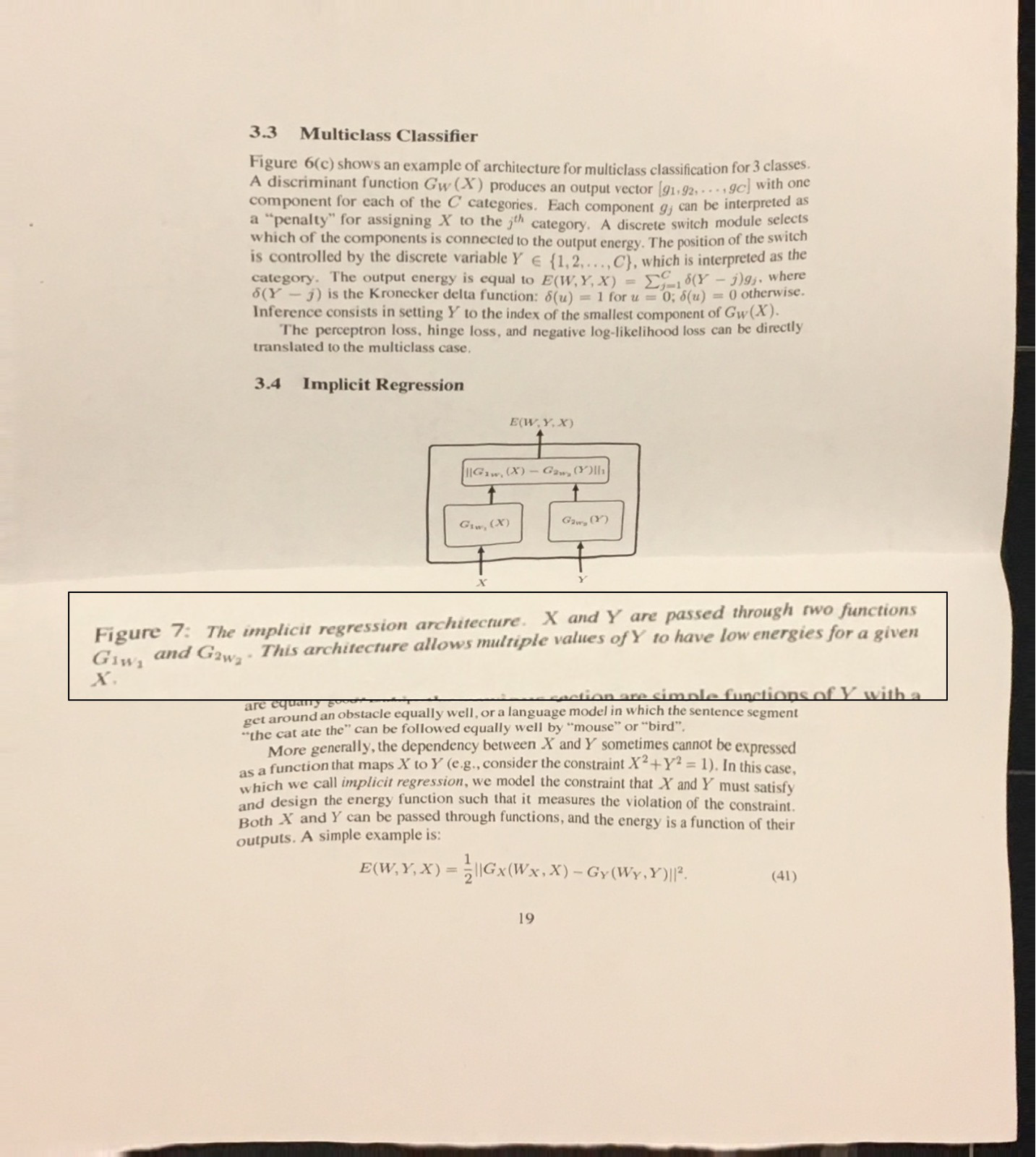} &
\includegraphics[width=0.28\linewidth]{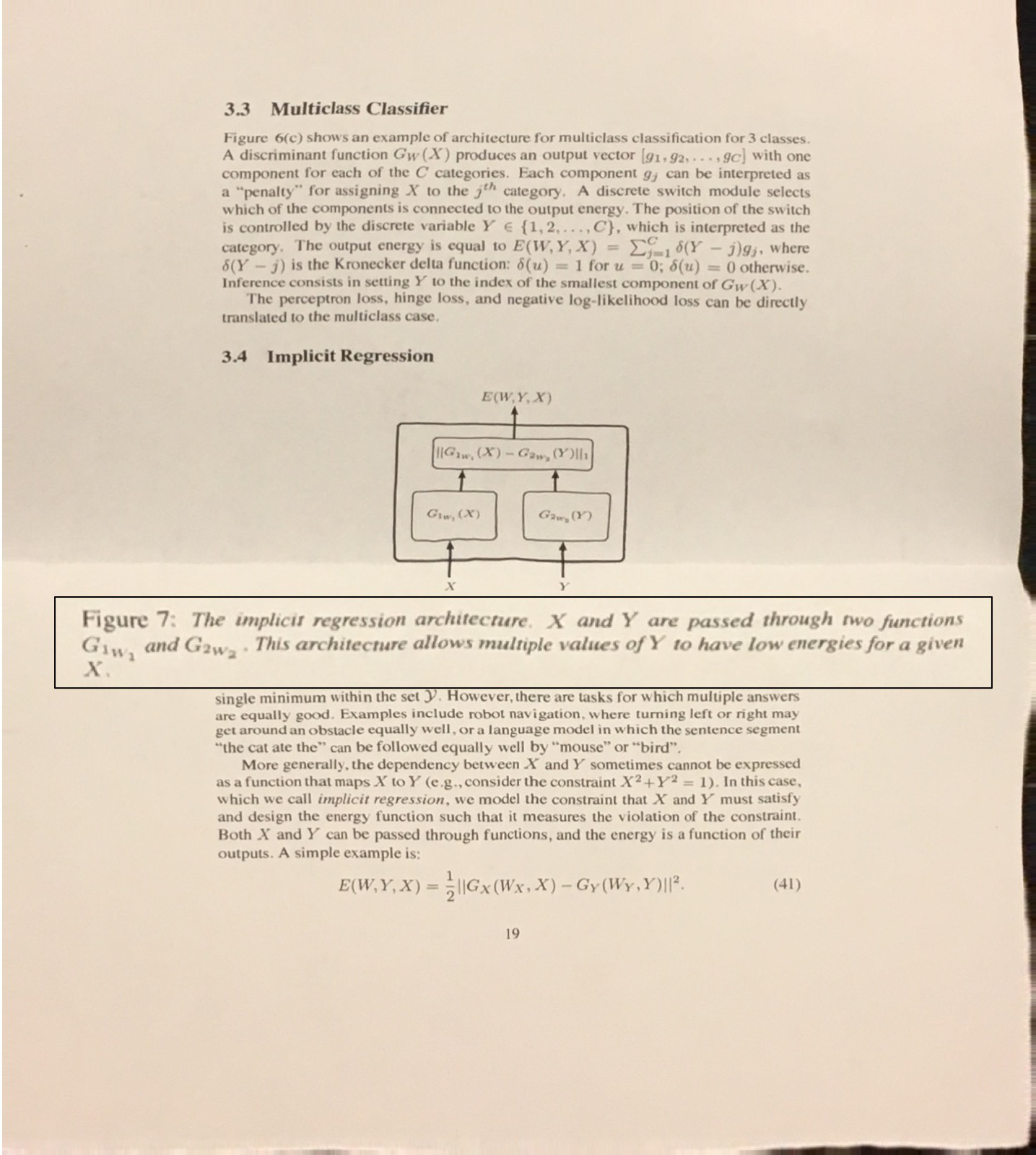} \\

\end{tabular}

\end{center}
\caption{\label{supp_fig:additional_results}
    {\small \textbf{Additional Results from the real image dataset of~\cite{ma2018docunet}.} 
}
\vspace{-0.2cm}}
\end{figure}

%% file: main.bbl
\begin{thebibliography}{10}
\providecommand{\url}[1]{\texttt{#1}}
\providecommand{\urlprefix}{URL }
\providecommand{\doi}[1]{https://doi.org/#1}

\bibitem{baek2019STRcomparisons}
Baek, J., Kim, G., Lee, J., Park, S., Han, D., Yun, S., Oh, S.J., Lee, H.: What
  is wrong with scene text recognition model comparisons? dataset and model
  analysis. In: International Conference on Computer Vision (ICCV) (2019), to
  appear

\bibitem{Ramanna2019deeplearning}
Bajjer~Ramanna, V.K., Bukhari, S.S., Dengel, A.: Document image dewarping using
  deep learning. In: The 8th International Conference on Pattern Recognition
  Applications and Methods. International Conference on Pattern Recognition
  Applications and Methods (ICPRAM-2019), February 19-21, Prague, Czech
  Republic. Insticc (2019)

\bibitem{Brown2004ImageRO}
Brown, M.S., Seales, W.B.: Image restoration of arbitrarily warped documents.
  IEEE Transactions on Pattern Analysis and Machine Intelligence  \textbf{26},
  1295--1306 (2004)

\bibitem{burden2019rectification}
Burden, A., Cote, M., Albu, A.B.: Rectification of camera-captured document
  images with mixed contents and varied layouts. In: 2019 16th Conference on
  Computer and Robot Vision (CRV). pp. 33--40. IEEE (2019)

\bibitem{blender}
Community, B.O.: Blender - a 3D modelling and rendering package. Blender
  Foundation, Stichting Blender Foundation, Amsterdam (2018),
  \url{http://www.blender.org}

\bibitem{das2019dewarpnet}
Das, S., Ma, K., Shu, Z., Samaras, D., Shilkrot, R.: Dewarpnet: Single-image
  document unwarping with stacked 3d and 2d regression networks. In: The IEEE
  International Conference on Computer Vision (ICCV) (October 2019)

\bibitem{das2017fold}
Das, S., Mishra, G., Sudharshana, A., Shilkrot, R.: The common fold: Utilizing
  the four-fold to dewarp printed documents from a single image. In:
  Proceedings of the 2017 ACM Symposium on Document Engineering. p. 125–128.
  DocEng ’17, Association for Computing Machinery, New York, NY, USA (2017).
  \doi{10.1145/3103010.3121030}, \url{https://doi.org/10.1145/3103010.3121030}

\bibitem{grning2018twostage}
Grüning, T., Leifert, G., Strauß, T., Michael, J., Labahn, R.: A two-stage
  method for text line detection in historical documents. International Journal
  on Document Analysis and Recognition (IJDAR) p. 285–302 (Jul 2018)

\bibitem{huang2017densely}
Huang, G., Liu, Z., Van Der~Maaten, L., Weinberger, K.Q.: Densely connected
  convolutional networks. In: Proceedings of the IEEE conference on computer
  vision and pattern recognition. pp. 4700--4708 (2017)

\bibitem{huang2015textline}
{Huang}, Z., {Gu}, J., {Meng}, G., {Pan}, C.: Text line extraction of curved
  document images using hybrid metric. In: 2015 3rd IAPR Asian Conference on
  Pattern Recognition (ACPR). pp. 251--255 (Nov 2015).
  \doi{10.1109/ACPR.2015.7486504}

\bibitem{textract}
Inc., A.: Amazon textract, \url{https://aws.amazon.com/textract}

\bibitem{gcp_ocr}
Inc., G.: Detect text in images, \url{https://cloud.google.com/vision/docs/ocr}

\bibitem{ma2018docunet}
Ke~Ma, Zhixin~Shu, X.B.J.W.D.S.: Docunet: Document image unwarping via a
  stacked u-net. In: Proceedings of IEEE Conference on Computer Vision and
  Pattern Recognition (2018)

\bibitem{kuhn1955hungarian}
Kuhn, H.W.: The hungarian method for the assignment problem. Naval research
  logistics quarterly  \textbf{2}(1-2),  83--97 (1955)

\bibitem{levenshtein1966binary}
Levenshtein, V.I.: Binary codes capable of correcting deletions, insertions,
  and reversals

\bibitem{li2019docrect}
Li, X., Zhang, B., Liao, J., Sander, P.V.: Document rectification and
  illumination correction using a patch-based cnn. ACM Transactions on Graphics
  (TOG)  (11 2019)

\bibitem{liu2018fots}
Liu, X., Liang, D., Yan, S., Chen, D., Qiao, Y., Yan, J.: Fots: Fast oriented
  text spotting with a unified network. 2018 IEEE/CVF Conference on Computer
  Vision and Pattern Recognition  (Jun 2018)

\bibitem{lyu2018masktextspotter}
Lyu, P., Liao, M., Yao, C., Wu, W., Bai, X.: Mask textspotter: An end-to-end
  trainable neural network for spotting text with arbitrary shapes. Lecture
  Notes in Computer Science p. 71–88 (2018)

\bibitem{ma2017arbitraryoriented}
Ma, J., Shao, W., Ye, H., Wang, L., Wang, H., Zheng, Y., Xue, X.:
  Arbitrary-oriented scene text detection via rotation proposals. IEEE
  Transactions on Multimedia  \textbf{20}(11),  3111–3122 (Nov 2018)

\bibitem{ren2015faster}
Ren, S., He, K., Girshick, R., Sun, J.: Faster r-cnn: Towards real-time object
  detection with region proposal networks. In: Advances in Neural Information
  Processing Systems 28, pp. 91--99. Curran Associates, Inc. (2015)

\bibitem{ronneberger2015u}
Ronneberger, O., Fischer, P., Brox, T.: U-net: Convolutional networks for
  biomedical image segmentation. In: International Conference on Medical image
  computing and computer-assisted intervention. pp. 234--241. Springer (2015)

\bibitem{smith2007tesseract}
Smith, R.: An overview of the tesseract ocr engine. In: Ninth International
  Conference on Document Analysis and Recognition (ICDAR 2007). vol.~2, pp.
  629--633. IEEE (2007)

\bibitem{Sorkine2005LaplacianMP}
Sorkine-Hornung, O.: Laplacian mesh processing. In: Eurographics (2005)

\bibitem{wang2004ssim}
Wang, Z., Bovik, A.C., Sheikh, H.R., Simoncelli, E.P.: Image quality
  assessment: from error visibility to structural similarity. IEEE transactions
  on image processing  \textbf{13}(4),  600--612 (2004)

\bibitem{You2016MultiviewRO}
You, S., Matsushita, Y., Sinha, S., Bou, Y.B., Ikeuchi, K.: Multiview
  rectification of folded documents. IEEE Transactions on Pattern Analysis and
  Machine Intelligence  \textbf{40},  505--511 (2016)

\bibitem{yousef2020origaminet}
Yousef, M., Bishop, T.E.: Origaminet: Weakly-supervised, segmentation-free,
  one-step, full page text recognition by learning to unfold. In: Proceedings
  of the IEEE/CVF Conference on Computer Vision and Pattern Recognition (2020)

\bibitem{zheng2014real}
Zheng, Y., Kang, X., Li, S., He, Y., Sun, J.: Real-time document image
  super-resolution by fast matting. In: 2014 11th IAPR International Workshop
  on Document Analysis Systems. pp. 232--236. IEEE (2014)

\bibitem{zhou2017east}
Zhou, X., Yao, C., Wen, H., Wang, Y., Zhou, S., He, W., Liang, J.: East: An
  efficient and accurate scene text detector. 2017 IEEE Conference on Computer
  Vision and Pattern Recognition (CVPR)  (Jul 2017)

\end{thebibliography}
